\documentclass[preprints,review,accept,pdftex,moreauthors]{Definitions/mdpi} 
\firstpage{1} 
\pubvolume{1}
\issuenum{1}
\articlenumber{0}
\pubyear{2025}
\copyrightyear{2025}
\datereceived{ } 
\daterevised{ } 
\dateaccepted{ } 
\datepublished{ } 
\hreflink{https://doi.org/} 

\Title{Deep Learning for Spatio-Temporal Fusion in Land Surface Temperature Estimation: A Comprehensive Survey, Experimental Analysis, and Future Trends}

\TitleCitation{Deep Learning for Spatio-Temporal Fusion in Land Surface Temperature Estimation: A Comprehensive Survey, Experimental Analysis, and Future Trends}

\Author{Sofiane Bouaziz $^{1,2}$\orcidA{}, Adel Hafiane $^{2,}$*\orcidB{}, Raphaël Canals $^{2}$\orcidC{}, Rachid Nedjai $^{3}$\orcidD{}}

\AuthorNames{Sofiane Bouaziz, Adel Hafiane, Raphaël Canals and Rachid Nedjai}

\AuthorCitation{Bouaziz, S.; Hafiane, A.; Canals, R.; Nedjai, R.}

\address{%
$^{1}$ \quad INSA CVL, Université d’Orléans, PRISME UR 4229, Bourges, 18022, Centre Val de Loire, France \\
$^{2}$ \quad Université d’Orléans, INSA CVL, PRISME UR 4229, Orléans, 45067, Centre Val de Loire, France \\
$^{3}$ \quad Université d'Orléans, CEDETE, UR 1210, Orléans, 45067, Centre Val de Loire, France}

\corres{Correspondence: adel.hafiane@univ-orleans.fr}

\abstract{Land Surface Temperature (LST) plays a key role in climate monitoring, urban heat assessment, and land-atmosphere interactions. However, current thermal infrared satellite sensors cannot simultaneously achieve high spatial and temporal resolution. Spatio-temporal fusion (STF) techniques address this limitation by combining complementary satellite data, one with high spatial but low temporal resolution, and another with high temporal but low spatial resolution. Existing STF techniques, from classical models to modern deep learning (DL) architectures, were primarily developed for surface reflectance (SR). Their application to thermal data remains limited and often overlooks LST-specific spatial and temporal variability. This study provides a focused review of DL-based STF methods for LST. We present a formal mathematical definition of the thermal fusion task, propose a refined taxonomy of relevant DL methods, and analyze the modifications required when adapting SR-oriented models to LST. To support reproducibility and benchmarking, we introduce a new dataset comprising 51 Terra MODIS-Landsat LST pairs from 2013 to 2024, and evaluate representative models to explore their behavior on thermal data. The analysis highlights performance gaps, architecture sensitivities, and open research challenges. The dataset and accompanying resources are publicly available at https://github.com/Sofianebouaziz1/STF-LST.}


\begin{document}


\section{Introduction}
\label{sec:introduction}
Over the past decade, urbanization has intensified globally, with cities accommodating an increasing proportion of the world's population. By 2020, more than half of people lived in urban areas, and European urbanization had already reached 74\% by 2017~\cite{population20072007}. This rapid growth intensifies environmental challenges, including the Urban Heat Island (UHI) effect, elevated air and water pollution, and reduced green spaces. Collectively, these impacts increase energy demand, degrade air quality, and threaten public health through higher rates of respiratory illnesses and heat-related mortality~\cite{vanos2015association, nuruzzaman2015urban, masiol2014thirteen}.

Land Surface Temperature (LST) is a key variable for understanding and managing these environmental processes. Physically, it represents the thermal radiation emitted by the Earth's surface, reflecting how incoming solar energy interacts with bare ground or vegetated canopies~\cite{HULLEY201957}. It plays a key role in revealing the temporal and spatial dynamics of the surface's equilibrium state~\cite{li2013relationship, kerr2004land}. This information supports a wide range of applications, including climate monitoring~\cite{schneider2010space, hall2012satellite}, urban planning~\cite{ibrahim2012land, maimaitiyiming2014effects}, and natural resource management~\cite{luyssaert2014land, kafy2020modelling}. Satellites provide the primary means of measuring LST at regional to global scales, offering consistent coverage and frequent revisit times~\cite{li2013satellite}. Consequently, many thermal infrared (TIR) sensors have been deployed, that led to diverse approaches for estimating LST at different spatial and temporal resolutions~\cite{wan1996generalized, gillespie1998temperature, sun2003estimation, trigo2011satellite, malakar2018operational, koetz2018high, shen2020generating}. Despite these advances, technical and budgetary constraints impose a fundamental trade-off between spatial and temporal resolutions~\cite{chen2015comparison, shen2016integrated}. Spatial resolution refers to the level of detail represented within a single pixel of a satellite image~\cite{ZHANG2023129}, while temporal resolution describes the frequency of observations over time for a given region~\cite{20201Shunlin}. Achieving both high spatial and high temporal resolutions remains difficult, as finer spatial detail typically reduces revisit frequency, whereas frequent acquisitions rely on coarser spatial coverage~\cite{zhan2013disaggregation}.

\begin{figure}[H]
\centering
\includegraphics[width=12cm]{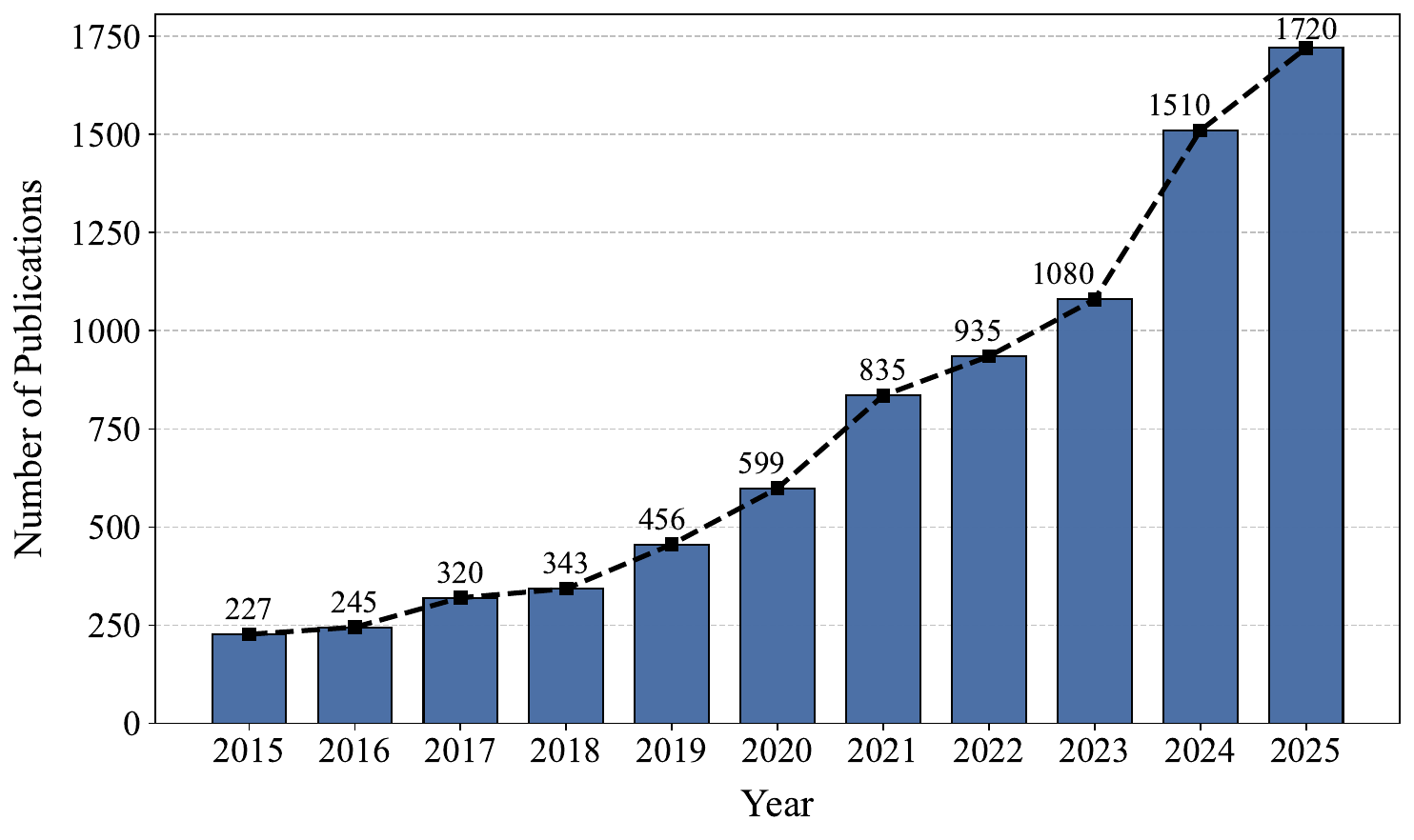}
\caption{Yearly literature count related to STF for LST estimation indexed by Google Scholar since 2015. The search query also covered the synonyms of STF, including data and image fusion.\label{fig:articles_per_year}}
\end{figure}

Two main strategies exist for producing LST data with both high spatial and temporal resolutions~\cite{mao2021resolution}: spatial downscaling and spatio-temporal fusion (STF)~\cite{belgiu2019spatiotemporal}. Spatial downscaling, also referred to as thermal sharpening or disaggregation, enhances spatial resolution by assuming a stable relationship between LST and auxiliary variables across scales~\cite{agam2007vegetation, nichol2009emissivity, duan2016spatial, xinming2021spatial}. For example, the traditional thermal sharpening approach~\cite{agam2007vegetation} estimates fine-resolution LST using predictors such as normalized difference vegetation index (NDVI) and fractional vegetation cover, and later studies introduced additional variables including the normalized difference building index (NDBI)~\cite{agathangelidis2019improving} and surface albedo~\cite{dominguez2011high}. However, this dependency limits robustness, as the relationship between LST and auxiliary features often changes over time~\cite{zhan2016disaggregation}. STF, also known as image or data fusion, offers an alternative strategy by combining two remote sensing (RS) satellite sources that share similar spectral characteristics but differ in spatial and temporal resolutions~\cite{zhu2018spatiotemporal}. In most cases, one sensor provides high spatial and low temporal resolutions (HSLT), while the other offers low spatial and high temporal resolutions (LSHT)~\cite{song2018spatiotemporal}. Figure \ref{fig:articles_per_year} illustrates the annual number of publications on STF for LST estimation from $2015$ to $2025$. The field has experienced rapid growth, rising from $227$ papers in $2015$ to $1720$ papers in $2025$. This surge reflects increasing interest in high spatial and high temporal (HSHT) LST research, driven by both advances in satellite sensors and the urgent need to address climate-related challenges.

Current STF approaches are commonly classified into four categories: weighted-based, unmixing-based, hybrid, and learning-based methods. Weighted-based methods produce fine-resolution satellite images by leveraging neighborhood information~\cite{gao2006blending, zhu2010enhanced, kim2012evaluation, wu2013land, huang2013generating, wu2015integrated}. For instance, STARFM~\cite{gao2006blending} adjusts pixel weights using separate coefficients for homogeneous and heterogeneous regions, and ESTARFM~\cite{zhu2010enhanced} further refines weights by considering mixed and pure pixels. These approaches are fast and stable, particularly in homogeneous areas, but their reliance on homogeneity assumptions limits detailed reconstruction. Unmixing-based approaches treat coarse-resolution temporal variations as mixtures of finer-scale components~\cite{wu2012use, zhang2013enhanced, zhang2015generalization, wu2015generating}, yet they often produce blocky artifacts due to limited temporal modeling. Hybrid methods integrate weighted-based and unmixing-based principles to exploit their respective strengths~\cite{zhu2016flexible, li2017generating, quan2018integrated, xia2019combining}. Learning-based methods rely on training models with existing datasets to learn mappings between HSLT and LSHT images. They are broadly divided into sparse representation-based and deep learning (DL)-based approaches. Sparse representation methods (or dictionary learning)~\cite{huang2012spatiotemporal, wu2015error, wei2016spatiotemporal, peng2021spatiotemporal} employ Bayesian learning to model coarse-to-fine satellite image relationships. Examples include covariance functions~\cite{li2013blending}, low-pass filtering~\cite{huang2013unified}, pixel unmixing~\cite{liao2016bayesian}, joint Gaussian distributions~\cite{xue2017bayesian}, and Kalman filters~\cite{addesso2014sequential}. These methods, however, depend strongly on satellite image range and involve complex computations~\cite{chen2022spatiotemporal}. DL-based approaches overcome these limitations by capturing non-linear relationships between input and output satellite images. Architectures explored for STF include Convolutional Neural Networks (CNNs)~\cite{song2018spatiotemporal, tan2018deriving, liu2019stfnet, zheng2019spatiotemporal, yin2020spatiotemporal, wang2020spatiotemporal, li2020new, qin2022mustfn, sun2023supervised, meng2022spatio, yu2024unsupervised}, AutoEncoders (AEs)~\cite{tan2019enhanced, chen2022spatiotemporal}, Generative Adversarial Networks (GANs)~\cite{chen2020cyclegan, zhang2020remote, ma2021explicit, tan2021flexible, zhang2021object, song2022mlff, tan2022robust, pan2023adaptive, huang2024stfdiff, chen2025cgmfn}, Vision Transformers (ViT)~\cite{li2021msnet, yang2022msfusion, chen2022swinstfm, li2022enhanced, jiang2024cnn, benzenati2024stf, ma2025sft, hu2025two}, and Recurrent Neural Networks (RNNs)~\cite{yang2021robust, zhan2024time, meng2022spatio}.

Although a substantial body of STF research exists, most methods were originally developed and validated using SR datasets, and transferring them directly to LST is nontrivial due to fundamentally different spatio-temporal dynamics. SR varies mainly with phenology and seasonal illumination changes, resulting in relatively smooth spatial transitions and limited short-term fluctuations. In contrast, LST exhibits rapid temporal variability, often at hourly or sub-hourly scales, driven by atmospheric conditions, surface energy balance, wind, cloud dynamics, and material-dependent thermal properties~\cite{prata1995thermal, wu2021spatially}. Spatially, LST can also change abruptly over short distances, such as between impervious surfaces and vegetation, whereas SR typically transitions more gradually with land cover patterns~\cite{prata1995thermal}. A number of surveys have examined STF in RS, but their coverage, technical depth, and relevance to LST vary considerably. Table~\ref{tab:comparative_surveys} compares existing surveys across seven criteria. The first is publication year, with more recent works prioritized for their coverage of the latest advancements. The second is the target domain, where, for LST-focused surveys, we assess whether they address how SR-based STF methods could be adapted for LST. We further evaluate whether surveys include DL methods, discuss their limitations, and highlight open challenges. Finally, we consider whether the surveys provide experimental evaluation and whether they introduce or benchmark a dataset to support future research. To date, only three surveys \cite{wu2021spatially, yoo2020spatial, ran2024review} explicitly address STF for LST. However, none provide a detailed examination of learning-based models, particularly DL, and none perform experimental comparison or challenge the assumption that SR-based STF techniques generalize directly to LST.

In this work, we review recent advances in DL-based STF for LST estimation. To summarize, our main contributions are the following:

\begin{itemize}
    \item We provide a comprehensive overview of DL-based STF methods for LST, highlighting their architectures, objectives, and adaptations for LST’s spatio-temporal dynamics.
    \item We introduce an open-source MODIS-Landsat LST pair dataset (STF-LST), comprising 51 images spanning 2013-2024, which serves as the first benchmark in the field.
    \item We conduct experimental analysis of state-of-the-art DL methods by offering quantitative and qualitative insights into their performance, limitations, and practical applicability for LST estimation.
\end{itemize}

\begin{table}[t]
\centering
\caption{Comparative analysis of existing STF surveys in RS. Criteria include: (1) Publication year, (2) Application scope (SR, NDVI, LST), (3) Adaptation of SR methods for LST, (4) Depth of DL coverage, (5) Discussion of DL limitations and open challenges, (6) Experimental evaluation, (7) Introduction of a benchmark dataset. A checkmark {\color{darkgreen}\cmark} indicates substantive coverage.}
\renewcommand{\arraystretch}{1}

\begin{adjustwidth}{-1.5cm}{-1.5cm}
\begin{tabularx}{\fulllength}{LCCCCCCC}
\toprule
\textbf{Survey} & \textbf{Year} & \textbf{Application scope} & \textbf{Adaptation to LST} & 
\textbf{Deep Learning} & \textbf{Open challenges} & \textbf{Experimental evaluation} & \textbf{New Dataset} \\
\midrule
\cite{zhu2018spatiotemporal} & 2018 & SR & {\color{darkred}\xmark} & {\color{darkred}\xmark} & {\color{darkred}\xmark} & {\color{darkred}\xmark} & {\color{darkred}\xmark} \\
\cite{belgiu2019spatiotemporal} & 2019 & SR & {\color{darkred}\xmark} & {\color{darkred}\xmark} & {\color{darkred}\xmark} & {\color{darkred}\xmark} & {\color{darkred}\xmark} \\
\cite{li2020spatio} & 2020 & SR & {\color{darkred}\xmark} & {\color{darkgreen}\cmark} & {\color{darkred}\xmark} & {\color{darkgreen}\cmark} & {\color{darkgreen}\cmark} \\
\cite{yoo2020spatial} & 2020 & LST & {\color{darkred}\xmark} & {\color{darkred}\xmark} & {\color{darkred}\xmark} & {\color{darkred}\xmark} & {\color{darkred}\xmark} \\
\cite{wang2020spatiotemporal} & 2020 & SR & {\color{darkred}\xmark} & {\color{darkgreen}\cmark} & {\color{darkgreen}\cmark} & {\color{darkgreen}\cmark} & {\color{darkred}\xmark} \\
\cite{wu2021spatially} & 2021 & LST & {\color{darkred}\xmark} & {\color{darkred}\xmark} & {\color{darkred}\xmark} & {\color{darkred}\xmark} & {\color{darkred}\xmark} \\
\cite{ferchichi2022forecasting} & 2022 & NDVI & {\color{darkred}\xmark} & {\color{darkgreen}\cmark} & {\color{darkgreen}\cmark} & {\color{darkred}\xmark} & {\color{darkred}\xmark} \\
\cite{wang2023review} & 2023 & SR & {\color{darkgreen}\cmark} & {\color{darkred}\xmark} & {\color{darkgreen}\cmark} & {\color{darkgreen}\cmark} & {\color{darkred}\xmark} \\
\cite{wang2023comprehensive} & 2023 & SR & {\color{darkred}\xmark} & {\color{darkred}\xmark} & {\color{darkred}\xmark} & {\color{darkred}\xmark} & {\color{darkred}\xmark} \\
\cite{xiao2023review} & 2023 & SR & {\color{darkred}\xmark} & {\color{darkgreen}\cmark} & {\color{darkgreen}\cmark} & {\color{darkred}\xmark} & {\color{darkred}\xmark} \\
\cite{chen2023spatiotemporal} & 2023 & SR & {\color{darkred}\xmark} & {\color{darkred}\xmark} & {\color{darkred}\xmark} & {\color{darkred}\xmark} & {\color{darkred}\xmark} \\
\cite{cui2024comprehensive} & 2024 & SR & {\color{darkred}\xmark} & {\color{darkred}\xmark} & {\color{darkred}\xmark} & {\color{darkred}\xmark} & {\color{darkred}\xmark} \\
\cite{anand2024pansharpening} & 2024 & SR & {\color{darkred}\xmark} & {\color{darkred}\xmark} & {\color{darkred}\xmark} & {\color{darkred}\xmark} & {\color{darkred}\xmark} \\
\cite{ran2024review} & 2024 & LST & {\color{darkgreen}\cmark} & {\color{darkred}\xmark} & {\color{darkred}\xmark} & {\color{darkred}\xmark} & {\color{darkred}\xmark} \\
\cite{swain2024spatio} & 2024 & SR & {\color{darkred}\xmark} & {\color{darkred}\xmark} & {\color{darkred}\xmark} & {\color{darkred}\xmark} & {\color{darkred}\xmark} \\
\cite{lian2025recent} & 2025 & SR & {\color{darkred}\xmark} & {\color{darkgreen}\cmark} & {\color{darkgreen}\cmark} & {\color{darkgreen}\cmark} & {\color{darkred}\xmark} \\
\cite{sun2025decade} & 2025 & SR & {\color{darkred}\xmark} & {\color{darkgreen}\cmark}  & {\color{darkgreen}\cmark}  & {\color{darkgreen}\cmark}  & {\color{darkred}\xmark} \\

\textbf{Ours} & 2025 & LST & {\color{darkgreen}\cmark} & {\color{darkgreen}\cmark} & {\color{darkgreen}\cmark} & {\color{darkgreen}\cmark} & {\color{darkgreen}\cmark} \\
\bottomrule
\end{tabularx}
\label{tab:comparative_surveys}
\end{adjustwidth}
\end{table}

This review is organized as follows. Section~\ref{sec:lst_extraction} introduces satellite-derived LST, highlighting its physical meaning, the trade-offs between spatial and temporal resolution, and its fundamental differences from SR. Section~\ref{sec:stf_formulation} formulates the STF problem for LST estimation, including its mathematical definition, commonly used loss functions, and evaluation metrics. In Section~\ref{sec:taxonomy}, we propose a novel taxonomy of DL-based STF methods. Section~\ref{sec:test} presents an extensive experimental analysis, where representative STF approaches are evaluated on paired MODIS-Landsat LST data. Section~\ref{sec:limitation} discusses the limitations of current methods and outlines promising future research directions. Finally, Section~\ref{sec:conclusion} summarizes the main findings of this survey.

\section{Satellite-Derived LST}
\label{sec:lst_extraction}
In this section, we will define LST, explain how it is derived from satellite observations, discuss the challenges associated with the trade-off between spatial and temporal resolution in satellite data, and highlight the differences between LST and SR. 

\subsection{LST Concept and Retrieval}
\label{sec:lstretrieval}
\begin{figure}[H]
\centering
\includegraphics[width=8cm]{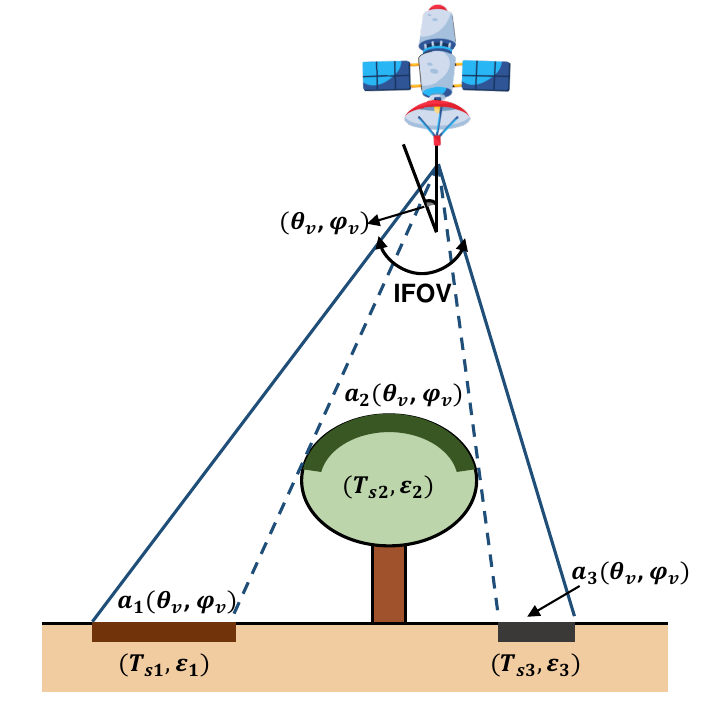}
\caption{Satellite-derived LST, inspired by~\cite{li2023satellite}. $T_{s i}$, $\varepsilon_i$, and $a_i$ represent the surface temperature, emissivity, and projected area weight for the $i$-th visible component, respectively. $\theta_v$ is the view zenith angle, and $\varphi_v$ is the viewing azimuth angle. \label{fig:Satellite-derived-LST}}
\end{figure}

LST is defined as the thermodynamic temperature at the surface of objects. In RS, this surface layer is continuous and projected across all visible components within the sensor’s instantaneous field of view (IFOV), as shown in Figure \ref{fig:Satellite-derived-LST}. LST is called radiometric temperature~\cite{norman1995terminology}, as it requires eliminating atmospheric effects and correcting for emissivity. Its formula is described in the Equation \ref{eq:temp_radiance} ~\cite{prata1995thermal, dash2002land, li2013satellite}.
\begin{equation}
\begin{aligned}
T_{\mathrm{s}}(\theta_v, \varphi_v)
&= B_\lambda^{-1}\left[\frac{A}{B}\right], \\[6pt]
A &= R_\lambda(\theta_v, \varphi_v)
- R_{at_\lambda \uparrow}(\theta_v, \varphi_v)
- \tau_\lambda(\theta_v, \varphi_v)\left(1 - \varepsilon_\lambda(\theta_v, \varphi_v)\right) R_{at_\lambda \downarrow}, \\[4pt]
B &= \tau_\lambda(\theta_v, \varphi_v)\,\varepsilon_\lambda(\theta_v, \varphi_v)
\end{aligned}
\label{eq:temp_radiance}
\end{equation}
\noindent where $T_{\mathrm{s}}$ is the radiometric temperature, $\theta_v$ and  $\varphi_v$ represent the viewing zenith and azimuth angles, $\lambda$ is the channel-effective wavelength. $B_\lambda^{-1}$ is the inverse function of Planck's law, which converts the measured radiance into temperature. $R_\lambda$, $R_{a t_\lambda \uparrow}$, and $R_{a t_\lambda \downarrow}$ are the at-sensor observed radiance, upward atmospheric radiance, and downward atmospheric radiance respectively, $\tau_\lambda$ is the channel atmospheric transmittance, and $\varepsilon_\lambda$ is the channel land surface emissivity (LSE).

The radiometric temperature has four properties:
\begin{enumerate}
    \item It is independent of spatial scale, meaning it can be applied across various scales~\cite{becker1995surface}.
    \item The depth of penetration depends on the wavelength used. In TIR range ($\lambda \approx 10 \, \mu\text{m}$), the depth varies from 1 to 100 $\mu\text{m}$, leading to the term skin temperature~\cite{norman1995terminology}. In the microwave range ($\lambda \approx 1 \, \text{cm}$), the depth ranges from 0.1 to 10 cm, hence the term subsurface temperature~\cite{duan2017cross}.
    \item It is affected by the viewing angle, making it directional~\cite{norman1995terminology}.
    \item It represents an average temperature derived from all homogeneous and isothermal elements within the IFOV~\cite{norman1995terminology}
\end{enumerate}

Retrieving LST from these radiances presents a challenging and ill-posed problem due to the following reasons:

\begin{itemize}
    \item LST retrieval is underdetermined because, for every radiance measurement across \( N \) TIR channels, there are \( N \) unknown LSEs and an unknown LST. 
    \item The radiances measured in different TIR channels are highly correlated, making the system of equations unstable and sensitive to small errors in the data. 
\end{itemize}

To overcome these challenges, additional assumptions and constraints are required to either increase the number of equations or reduce the number of unknowns, while decorrelating the data. Consequently, several algorithms have been developed to retrieve LST, including the Single-Channel~\cite{jimenez2014land}, Split-Window~\cite{rozenstein2014derivation}, and mono-window~\cite{wang2015improved}.

\subsection{Trade-offs in Spatial and Temporal Resolution for LST Retrieval}

 The accuracy of LST retrieval is significantly influenced by two key factors: \textit{spatial resolution} and \textit{temporal resolution}. 

 \begin{itemize}
    \item Spatial Resolution: defines the size of a pixel in the satellite image, which determines the smallest detectable feature. This is crucial for accurate LST retrieval, as fine-scale spatial resolution captures smaller, more localized temperature variations~\cite{20201Shunlin}.
    
    \item Temporal Resolution: refers to the frequency at which a satellite revisits the same area. A higher temporal resolution is critical for monitoring dynamic temperature changes over time~\cite{zhan2013disaggregation}.
\end{itemize}

Table \ref{tab:lstsat} summarizes the most commonly used satellites for LST retrieval, emphasizing the trade-off between spatial and temporal resolution. For instance, Landsat $8$ and $9$ provide high spatial resolution ($30$ m) but revisit the same area only every $16$ days. In contrast, MODIS Aqua and Terra offer daily observations ($1$ day temporal resolution) but at coarser spatial resolution ($1$ km). STF techniques present a promising approach to overcome this trade-off, and will be discussed in the following sections.

\begin{table}[H]
\caption{Comparison of common satellite-based LST products with their thermal Sensors, spatial and temporal resolutions, and temporal extents.\label{tab:lstsat}}
\begin{adjustwidth}{-1.5cm}{-1.5cm}
\renewcommand{\arraystretch}{1}
\begin{tabularx}{\fulllength}{l@{\hspace{-10pt}}C@{\hspace{-10pt}}C@{\hspace{-10pt}}C@{\hspace{-10pt}}C}
\toprule
\textbf{Satellite} & \textbf{Thermal Sensor} & \textbf{Spatial Resolution} & \textbf{Temporal Resolution} & \textbf{Temporal Extent} \\
\midrule

GF-5        & VIMS      & 40 m                        & 7 days          & 2018/05/09 -- Present \\
Landsat 9   & TIRS-2    & 100 m, resampled to 30 m     & 16 days         & 2021/09/27 -- Present \\
Landsat 8   & TIRS      & 100 m, resampled to 30 m     & 16 days         & 2013/02/11 -- Present \\
Landsat 7   & ETM+      & 100 m, resampled to 30 m     & 16 days         & 1999/04/15 -- Present (Partially) \\
Landsat 5   & TM        & 120 m, resampled to 30 m     & 16 days         & 1984/05/01 -- 2013/06/05 \\
Terra       & ASTER     & 90 m                         & 16 days         & 1999/12/18 -- Present \\
Aqua        & MODIS     & 1 km                         & 1 day           & 2002/05/04 -- Present \\
Terra       & MODIS     & 1 km                         & 1 day           & 1999/12/18 -- Present \\
Sentinel-3A & SLSTR     & 1 km                         & 1 day           & 2016/02/16 -- Present \\
FY-3D       & MERSI-2   & 375 m                        & 12 hours        & 2017/11/15 -- Present \\
SNPP        & VIIRS     & 375 m                        & 12 hours        & 2011/11/28 -- Present \\
GOES-8      & Imager    & 4 km                         & 30 minutes      & 1994/09/16 -- 2009/06/04 \\
FY-2F       & VISSR     & 5 km                         & 1 hour          & 2012/01/13 -- Present \\

\bottomrule
\end{tabularx}
\end{adjustwidth}
\end{table}

\subsection{Differences Between SR and LST dynamics}

SR and LST are both derived from satellite observations, but they display fundamentally different spatial and temporal behaviors because they are governed by distinct physical processes. This shapes how each variable changes across space and time.

\subsubsection{Spatial Variations} SR exhibits relatively smooth, structured, and stable spatial patterns~\cite{gomez2016optical}. For instance, forests, urban areas, and water bodies generally maintain consistent boundaries over time. SR values also change gradually across space, with neighboring pixels usually showing similar values. In contrast, LST can vary abruptly over short distances~\cite{li2013satellite}. For example, within a single urban area, surfaces with similar reflectance can exhibit large LST differences due to variations in thermal inertia. 

\subsubsection{Temporal Variations}

SR changes gradually over time, driven by seasonal or phenological processes such as vegetation growth, senescence, crop cycles, and land cover changes. As a result, SR exhibits strong temporal correlation and relatively predictable trends~\cite{chraibi2022stability}. LST, however, is subject to rapid temporal fluctuations due to diurnal cycles, precipitation events, wind, and other atmospheric conditions~\cite{li2013satellite}. LST can also vary significantly within hours, making it much less temporally stable than SR.

\section{STF problem formulation for LST}
\label{sec:stf_formulation}
This section presents a mathematical formulation of the STF problem for LST estimation. We introduce the relevant notations, describe how DL models are incorporated into the fusion process, categorize the loss functions commonly used during training, and outline the evaluation metrics employed to assess model performance.

\subsection{Mathematical Definition}
\label{sec:mathematicalformulation}

The STF problem can be formulated as a multi-objective optimization task, aiming to simultaneously improve both the spatial and temporal resolutions of LST data. Formally, this can be expressed as in Equation \ref{eq:multiobj}. 
\begin{equation}
\max \left( \mathcal{F}(R_s), \mathcal{F}(R_t) \right),
\label{eq:multiobj}
\end{equation}
\noindent where \( \mathcal{F}(R_s) \) and \( \mathcal{F}(R_t) \) are objective functions corresponding to spatial and temporal resolutions enhancement, respectively. Specifically, \( \mathcal{F}(R_s) \) aims to maximize spatial fidelity by enhancing the resolution and preserving fine-scale details in the high-resolution satellite image, while \( \mathcal{F}(R_t) \) seeks to maximize temporal fidelity by minimizing the temporal gap between consecutive acquisitions. An additional, often overlooked, objective is gap filling in the high-resolution satellite image, making the STF problem a tri-objective optimization task: balancing spatial and temporal resolutions enhancement along with accurate reconstruction of missing or masked pixels. STF methods leverage known patterns in satellite image pixel values over time (temporal variations) and across spatial scales (spatial variations) to estimate HSHT satellite images. For LST, temporal variations refer to changes in pixel values observed at the same location over time, while spatial variations refer to differences in pixel values between LSHT and HSLT images.

\begin{figure}[h]
\begin{adjustwidth}{-1.5cm}{-1.5cm}
\centering
\includegraphics[width=16cm]{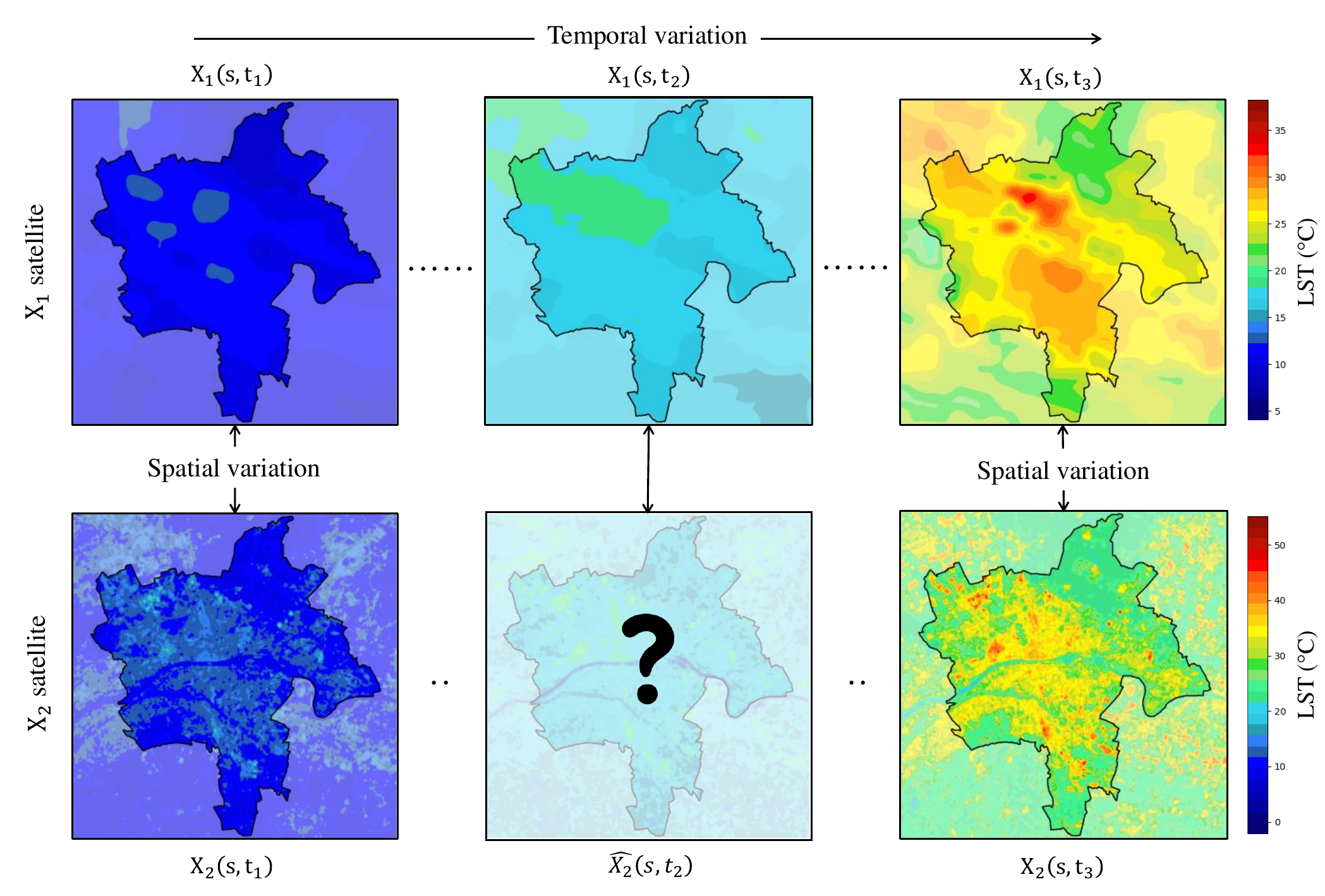} 
\end{adjustwidth}
\caption{Graphic representation of the STF for LST estimation. $X_1$ denotes data from the MODIS Terra satellite, and $X_2$ refers to data from the Landsat 8 satellite. The ROI, $s$, corresponds to Orléans Métropole, in France. The time steps $t_1$, $t_2$, and $t_3$ represent 6 Mar, 22 Mar, and 9 May 2022, respectively. \label{fig:STF-LST_rep}}
\end{figure}

Let \( X_1 \) and \( X_2 \) denote data from two satellites: \( X_1 \) provides LSHT LST data, while \( X_2 \) provides HSLT LST data. Let \( t_1, t_2, t_3 \) be three distinct time steps, and let \( s \) denote the geographic region of interest (ROI). Then, \( X_1(s, t_i) \) and \( X_2(s, t_i) \) for \( i \in \{1, 2, 3\} \) represent the LST data from the first or second satellite at time \( t_i \) for location \( s \). Given two pairs of satellite LST images, \( P_1 = \{X_1(s, t_1), X_2(s, t_1)\} \) and \( P_3 = \{X_1(s, t_3), X_2(s, t_3)\} \), at times \( t_1 \) and \( t_3 \), along with the LSHT LST image \( X_1(s, t_2) \) at time \( t_2 \), the goal of STF is to predict the corresponding HSLT LST image \( X_2(s, t_2) \), as illustrated in Figure \ref{fig:STF-LST_rep}. Therefore, the predicted HSHT LST image at time \( t_2 \), denoted as \( \hat{X}_2(s, t_2) \), can be expressed as in Equation \ref{eq:stf_problem}.
\begin{equation}
\hat{X}_2(s, t_2) = f\left(P_1, P_3, X_1(s, t_2)\right),
\label{eq:stf_problem}
\end{equation}

\noindent where the objective is for \( \hat{X}_2(s, t_2) \) to approximate the true HSLT image \( X_2(s, t_2) \) as closely as possible. Table \ref{tab:notation} summarizes the notations used in this formulation.

\begin{table}[H]
\centering
\caption{Notations used in STF for LST estimation problem formulation.}
\renewcommand{\arraystretch}{1} 
\begin{tabular}{lc}
\toprule
\textbf{Notation} & \textbf{Significance} \\
\midrule
$X_1$ & Satellite providing LST data with LSHT \\
$X_2$ & Satellite providing LST data with HSLT \\
$t_i$ & Temporal time steps \\
$s$ & Region of interest \\
$X_i(s, t_i)$ & LST data at time $t_i$ for the region $s$ \\
$\hat{X}_2(s, t_i)$ & Predicted HSHT LST data at time $t_i$ and for region $s$ \\
$P_i$ & A pair of $X_1$ and $X_2$ for a specific ROI $s$ at time $t_i$ \\
\bottomrule
\end{tabular}
\label{tab:notation}
\end{table}

However, the STF problem cannot be effectively solved using linear methods~\cite{tran2017characterizing, zhang2015generalization, wang2021downscaling}. Non-linear approaches, such as DL, are better suited to capture the complex dependencies between LSHT and HSLT images. DL is a subset of machine learning that employs multi-layer neural networks to learn complex, non-linear relationships from data. To account for these non-linear dependencies, the function \( f \) is parameterized by a set of weights \( \mathbf{W} \), which are optimized during training, as shown in Equation \ref{eq:stf_dl}.
 \begin{equation}
\hat{X}_2(s, t_2) = f\left(P_1, P_3, X_1(s, t_2) \mid \mathbf{W}\right).
\label{eq:stf_dl}
\end{equation}

In practice, relying on two pairs of images may not always be feasible, as it requires waiting for a future pair, \( P_3 \), captured under favorable conditions (e.g., minimal cloud coverage) to predict the high-spatial LST at an earlier time. To address this limitation, two alternative strategies are commonly employed: (i) using only one pair of satellite images, \( P_1 \), or (ii) leveraging a time series of previous pairs. 

\subsection{Loss Functions}

The weights \( \mathbf{W} \) in Equation \ref{eq:stf_dl} are optimized by minimizing a loss function, which measures the discrepancy between the predicted and true high-resolution satellite image. Although many losses originate from general computer vision tasks, their role in STF is tied to preserving temperature gradients, spatial structure, and radiometric consistency.

In practice, the objective is expressed as a weighted combination of multiple complementary loss terms, as shown in Equation~\ref{eq:total_loss}, where each $\lambda_i$ represents the relative importance of its corresponding loss term. 
\begin{equation}
\begin{split}
\mathcal{L}_{total} = \lambda_1 \mathcal{L}_{content} + \lambda_2 \mathcal{L}_{vision} + \lambda_3 \mathcal{L}_{feature} + \lambda_4 \mathcal{L}_{Spectral} + \lambda_5 \mathcal{L}_{GAN},
\text{where} \sum_{i=1}^{5} \lambda_i = 1.
\end{split}
\label{eq:total_loss}
\end{equation}
\noindent Here, $\mathcal{L}_{content}$, $\mathcal{L}_{vision}$, $\mathcal{L}_{feature}$, $\mathcal{L}_{spectral}$, and $\mathcal{L}_{GAN}$ denote the content, vision, feature, spectral, and adversarial loss terms, respectively, which are detailed in the following subsections. During training, the loss is minimized using gradient-based optimization algorithms, most commonly Stochastic Gradient Descent (SGD) and Adam~\cite{kingma2014adam}.

\subsubsection{Content Loss}
It preserves radiometric consistency by ensuring that pixel values remain physically plausible and aligned with real LST values. The most common form is the Mean Squared Error (MSE), expressed in Equation \ref{eq:mse}, while variants such as Mean Absolute Error(MAE), KL divergence~\cite{tian2022recent}, Huber loss~\cite{huber1992robust}, or index-driven terms (e.g., NDVI-based~\cite{huang2021commentary} and NDBI-based~\cite{zha2003use} error constraints) have also been used to encode domain knowledge.
\begin{equation}
\mathcal{L}_{\text{content}} = \frac{1}{N} \sum_{i=1}^{N} \left( \hat{X}_2(s_i, t_2) - X_2(s_i, t_2) \right)^2
\label{eq:mse}
\end{equation}

\subsubsection{Vision Loss}
It penalizes visually unrealistic reconstructions and helps preserve spatial structure, reducing the risk of overly smoothed predictions that blur LST boundaries. It typically relies on perceptual similarity metrics such as Structural Similarity Index (SSIM) and MS-SSIM~\cite{wang2004image, zhao2016loss}, given by Equation \ref{eq:ssim}, which evaluate local luminance, contrast, and structural consistency rather than pixel-wise differences alone. It may also incorporate edge-aware formulations, such as Sobel-based loss~\cite{khare2021analysis}, to further enforce gradient continuity and retain sharp transitions.
\begin{equation}
\mathcal{L}_{\text{vision}} = 
1 - \text{SSIM}(\hat{X}_2(s,t_2), X_2(s,t_2))
\label{eq:ssim}
\end{equation}

\subsubsection{Feature Loss}
\label{sec:feature}
It captures high-level perceptual features by comparing intermediate representations extracted from pre-trained networks (e.g., encoders)~\cite{wu2017srpgan}. For LST, this loss helps preserve meaningful spatial patterns (e.g., thermal gradients between vegetation, water, and built-up areas) that may not be fully enforced through pixel-wise or vision losses alone. The feature loss is computed as the difference between the feature representations of the predicted image \( F_{\hat{X}_2} \) and the reference image \( F_{X_2} \), as expressed in Equation~\ref{eq:featureloss}, where \( L \) denotes the number of extracted feature elements.
\begin{equation}
\mathcal{L}_{\text{feature}} = \frac{1}{L} \| F_{\hat{X}_2} - F_{X_2} \|_2^2
\label{eq:featureloss}
\end{equation}

\subsubsection{Spectral Loss} 
\label{sec:Spectral}
It enforces consistency in the learned feature space by measuring the cosine similarity between the predicted and reference images~\cite{ma2021explicit}. In LST-based STF, this loss helps preserve thermal response patterns even when pixel-wise intensities vary (e.g., hotter built-up surfaces compared to cooler vegetation). The formulation is given in Equation~\ref{eq:spectralloss}.
\begin{equation}
\mathcal{L}_{\text{spectral}} = 1 - \frac{F_{\hat X_2} \cdot F_{X_2}}{\|F_{\hat X_2}\| \, \|F_{X_2}\|}
\label{eq:spectralloss}
\end{equation}

\subsubsection{Adversarial Loss}
It is used when the STF model relies on generative adversarial learning (e.g., GAN-based STF). The loss encourages the predicted high-resolution LST image to follow the distribution of real observation, making the output statistically indistinguishable from true high-resolution LST image~\cite{goodfellow2014generative}. 

\subsection{Evaluation Metrics}
\label{sec:evaluation}
STF models are typically evaluated using a diverse set of quantitative and qualitative metrics. These metrics can be grouped into three categories: Error Assessment, Quality Assessment, and Efficiency.

\begin{itemize}

\item \textit{Error Assessment Metrics:} These metrics measure the numerical discrepancy between the fused LST and the reference high-resolution one. Common examples include RMSE, MAE, relative MAE (rMAE), and the coefficient of determination (\(R^2\)). ERGAS is also frequently used to assess global normalized error, with lower values indicate higher fidelity.

\item \textit{Quality Assessment Metrics:} Rather than measuring numerical differences, this category assesses perceptual or structural similarity. SSIM~\cite{wang2004image}, Peak Signal-to-Noise Ratio (PSNR)~\cite{korhonen2012peak}, correlation coefficient (CC)~\cite{kaneko2003using}, Spectral Angle Mapper (SAM)~\cite{yuhas1992discrimination}, Perceptual Image Patch Similarity (LPIPS)~\cite{zhang2018unreasonable}, and Universal Image Quality Index (UIQI)~\cite{wang2002universal} are commonly adopted to quantify texture preservation, sharpness, and spectral consistency in reconstructed LST fields.

\item \textit{Efficiency Metrics:} Beyond accuracy, computational efficiency is increasingly emphasized, especially for large-scale or near real-time applications. Metrics include inference time (per fused scene), memory footprint, and scalability with spatial or temporal input size. For example, DL-based STF models are reported to achieve inference speeds orders of magnitude faster than classical algorithms such as ESTARFM~\cite{zhang2020remote}.

\end{itemize}

\section{Taxonomy of DL-based STF methods}
\label{sec:taxonomy}
\begin{figure}[h]
\centering
\includegraphics[width=12cm]{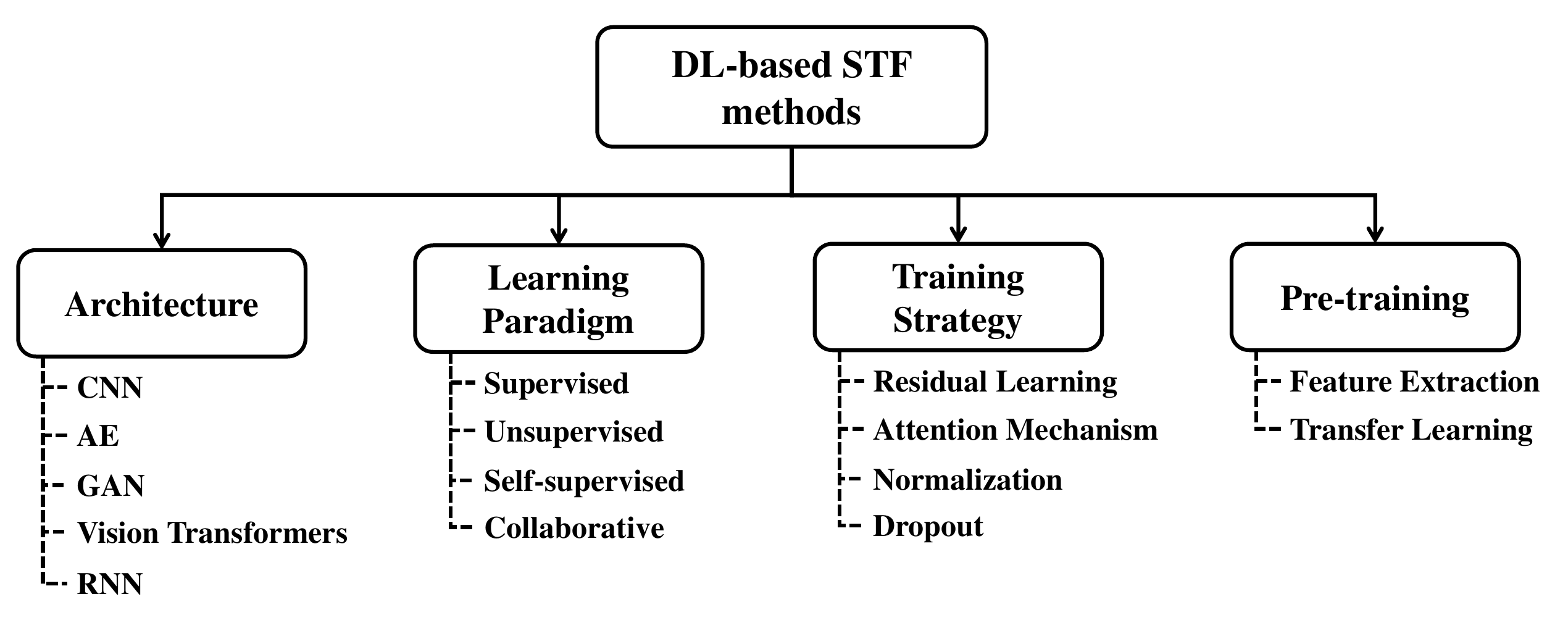}
\caption{Proposed Taxonomy for DL-Based STF Methods based on Architecture, Learning Paradigm, Training Strategy, and Incorporation of Pre-trained Models. Methods originally developed for related tasks (e.g., SR, NDVI) are included to highlight transferable design patterns relevant to LST estimation. \label{fig:STF-DL}}
\end{figure}

In this section, we present a taxonomy of DL-based STF methods constructed from $34$ representative works selected for recency, relevance, and methodological diversity. The classification is organized around four criteria: \textit{Architecture}, \textit{Learning Paradigm}, \textit{Training Strategy}, and \textit{Use of Pre-trained Models} (Figure~\ref{fig:STF-DL}). We adopt a broader perspective by including STF methods developed for related RS tasks, such as SR, NDVI, and others. We then map LST-specific methods onto this framework and discuss how each category can be adapted to address the unique spatial and temporal characteristics of LST data. This approach not only highlights shared methodological patterns but also clarifies their relevance and potential adaptation for LST estimation.

\subsection{Architectures}
STF methods are generally categorized based on their model architecture, a trend that also holds for STF-based LST estimation. Common architectures include \textit{CNNs}, \textit{AEs}, \textit{GANs}, \textit{ViT}, and \textit{RNNs}. 

\subsubsection{Convolutional Neural Networks}
CNNs are widely used architectures for processing grid-like data such as satellite images~\cite{krizhevsky2012imagenet, lecun2015deep}. They automatically learn spatial features from input data, making them effective for capturing complex patterns, edges, and textures. A typical CNN consists of layers such as \textit{convolutional}, \textit{pooling}, \textit{normalization}, \textit{activation}, and \textit{fully connected layers}~\cite{alzubaidi2021review}:

\begin{itemize}
    \item \textit{Convolutional Layers}: Extract spatial features using local filters.
    \item \textit{Pooling Layers}: Reduce spatial dimensions and improve feature robustness.
    \item \textit{Normalization Layers}: Stabilize and accelerate training.
    \item \textit{Activation Layers}: Introduce non-linearity (e.g., ReLU~\cite{nair2010rectified}).
    \item \textit{Fully Connected Layers}: Integrate learned features for prediction.
\end{itemize}

\begin{figure}[H]
\begin{adjustwidth}{-1.5cm}{-1.5cm}
\centering
\includegraphics[width=18cm]{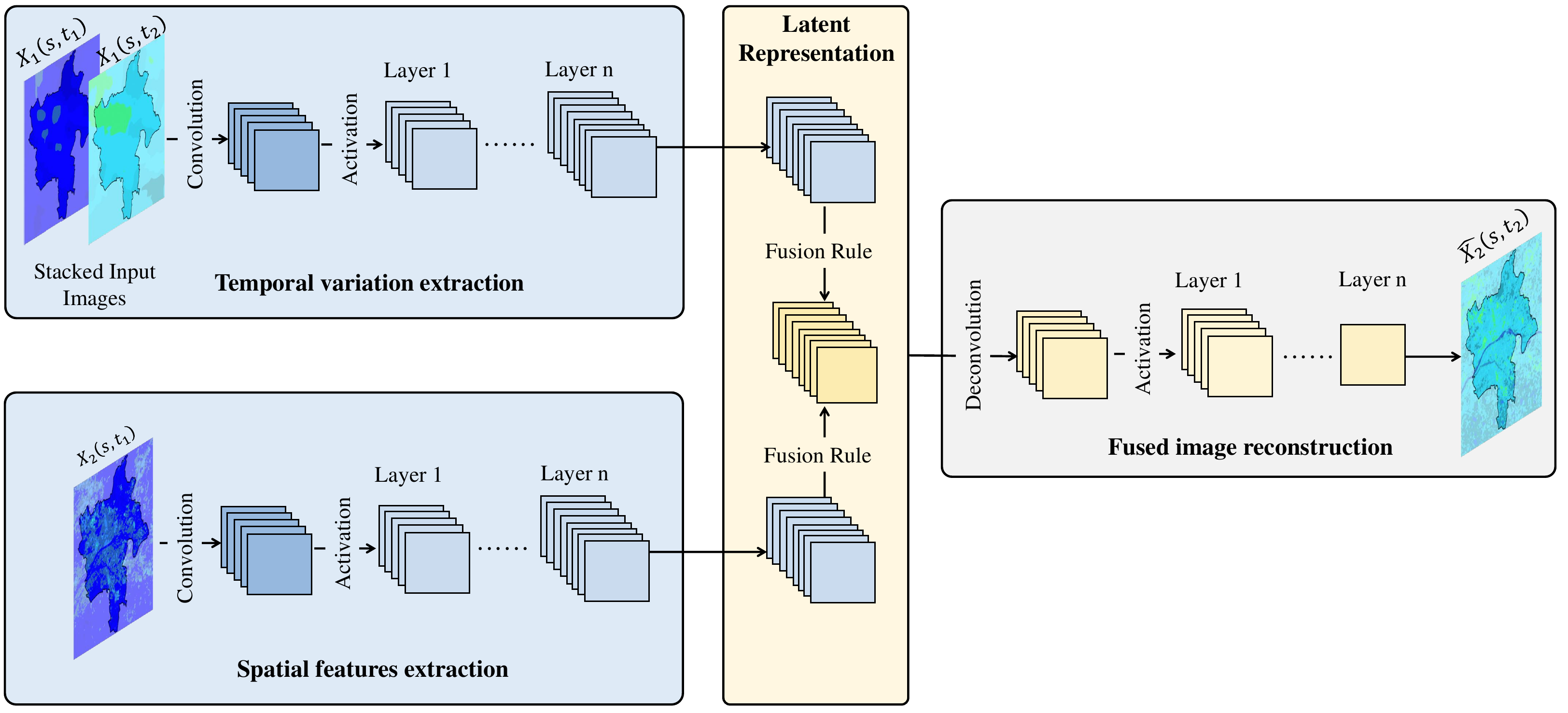} 
\end{adjustwidth}
\caption{Typical architecture of a CNN-based STF method using a single pair of images, $P_1$, composed of four main blocks: spatial feature extraction, temporal variation extraction, fusion of spatial and temporal representations, and satellite image reconstruction. The fusion rule can be element-wise addition, multiplication, concatenation, attention, etc. \label{fig:STF-CNN}}
\end{figure}

 CNN-based STF methods leverage CNNs to automatically model the complex, non-linear relationships between LSHT and HSLT satellite image pairs, which are then used to predict high-resolution target satellite images. Although CNNs are primarily designed for spatial modeling, temporal information can be incorporated by processing multiple temporal images through separate CNN streams and fusing their outputs to capture spatio-temporal variations. Figure~\ref{fig:STF-CNN} illustrates a typical CNN-based STF architecture using a single pair $P_1$. The network consists of four main blocks: (1) \textit{Spatial feature extraction}, where the high-resolution satellite image is processed through convolutional layers to obtain compressed spatial features, (2) \textit{Temporal variation extraction}, where two low-resolution satellite images are concatenated along channels and passed through convolutional layers to capture temporal features, (3) \textit{Feature fusion}, where spatial and temporal representations are combined in a latent space, and (4) \textit{Reconstruction}, where deconvolutional layers and fully connected layers generate the predicted high-resolution satellite image. When multiple pairs are used, each pair is processed similarly, and their weighted outputs are integrated to produce the final prediction. Table \ref{tab:CNN-STF} presents an overview of CNN-based DL methods for STF. The first CNN-based approach was introduced by \cite{song2018spatiotemporal}, where a CNN is trained on \( n \) MODIS-Landsat pairs and applied using \( P_1 \) and \( P_3 \) satellite pairs to predict the HSHT satellite image at the target date. The model employs an MSE loss and was evaluated on the CIA and LGC datasets. A similar strategy was adopted in \cite{zheng2019spatiotemporal}, while Wang et al. \cite{wang2020spatiotemporal} extended the formulation by incorporating two prior MODIS-Landsat pairs and using the Euclidean distance as the loss function. Tan et al. \cite{tan2018deriving} reformulated the problem according to the notation in Section~\ref{sec:mathematicalformulation}. Their method uses only one pair \( P_1 \) together with \( X_1(s,t_2) \) to estimate \( \hat{X}(s,t_2) \). Spatial features from Landsat at \( t_1 \) and temporal variations between the two MODIS dates are extracted through separate convolutional blocks, fused in a latent space, and reconstructed via deconvolution and fully connected layers. MSE is used as the loss, and the dataset is manually curated. Li et al. \cite{li2020new} proposed a sensor-bias-driven model that explicitly accounts for spectral and spatial inconsistencies between MODIS and Landsat. Two CNNs are used, one enhances coarse reflectance, while the second models cross-sensor bias. Qin et al. \cite{qin2022mustfn} introduced a multi-constrained loss that simultaneously improves fusion quality and enables gap filling, where the model learns to handle cloud-induced missing pixels through a binary mask mechanism. Yu et al. \cite{yu2024unsupervised} presented an unsupervised CNN framework for Landsat 8 and Sentinel-2 STF. Meng et al. \cite{meng2022spatio} combined a multiscale Siamese CNN with a convolutional RNN to jointly model spatial-spectral features and temporal dynamics. 

\begin{table}[H]
\caption{A comparative overview of CNN-based DL methods for STF. For each method, the table lists the satellite sensors used (\textit{Satellites}), whether the method is evaluated on LST data (\textit{LST}), the number of temporal pairs required for training (\textit{Pairs}), the loss functions employed (\textit{Loss Functions}), the evaluation metrics reported (\textit{Evaluation metrics}), the type of datasets used (\textit{Datasets}), and the availability of implementation code (\textit{Code}).}
\begin{adjustwidth}{-1.5cm}{-1.5cm}
\renewcommand{\arraystretch}{1}
\begin{tabularx}{\fulllength}{l@{\hspace{-1pt}}>{\centering\arraybackslash}m{3cm}@{\hspace{-12pt}}C@{\hspace{-30pt}}C@{\hspace{-15pt}}C >{\centering\arraybackslash}m{3.5cm} C@{\hspace{-10pt}}C@{\hspace{-10pt}}}
\toprule
\textbf{Method} & \textbf{Satellites} & \textbf{LST } & \textbf{Pairs} & \textbf{Loss Functions} & \textbf{Evaluation metrics} & \textbf{Datasets} & \textbf{Code} \\
\midrule

~\cite{song2018spatiotemporal} & MODIS, Landsat & \xmark & 2 & Content & RMSE, ERGAS, SAM, SSIM & CIA, LGC & \xmark \\

~\cite{tan2018deriving} & MODIS, Landsat & \xmark & 1 & Content & RMSE, $R^2$, SSIM & Hand-crafted & \href{https://github.com/theonegis/rs-data-fusion}{\cmark} \\

~\cite{liu2019stfnet} & MODIS, Landsat & \xmark & 2 & Content & RMSE, CC, SSIM & Hand-crafted & \xmark \\

~\cite{zheng2019spatiotemporal} & MODIS, Landsat & \xmark & 2 & Content & RMSE, SAM, SSIM, ERGAS & CIA, LGC & \xmark \\

~\cite{yin2020spatiotemporal} & MODIS, Landsat & \cmark & 2 & Content & RMSE, SSIM & Hand-crafted & \xmark \\

~\cite{wang2020spatiotemporal} & MODIS, Landsat & \xmark & 2 & Content & RMSE, CC, UIQI & Hand-crafted & \xmark \\

~\cite{li2020new} & MODIS, Landsat & \xmark & 2 & Content &RMSE, CC, ERGAS, SSIM, SAM& CIA, LGC &\xmark \\

~\cite{qin2022mustfn} & MODIS, Landsat & \xmark & 2 & Content, Vision &\begin{tabular}[t]{@{}c@{}} RMSE, $R^2$, MAE,\\ rMAE, MAEC\end{tabular}& Hand-crafted & \href{https://github.com/qpyeah/MUSTFN}{\cmark} \\

~\cite{meng2022spatio} & MODIS, Landsat & \xmark & 2 & Content, Vision & RMSE, SAM, SSIM & CIA, LGC & \xmark \\

~\cite{sun2023supervised} & \begin{tabular}[t]{@{}c@{}} MODIS, Landsat\\ Landsat, Sentinel-2 \end{tabular} & \cmark & $n$ & Content, Adversarial &RMSE, SSIM, ERGAS, PSNR, SAM& Hand-crafted & \xmark \\

~\cite{yu2024unsupervised} & Landsat, Sentinel-2 & \xmark & 1 & Content & RMSE, CC, SSIM & Hand-crafted & \xmark \\

\bottomrule
\end{tabularx}
\end{adjustwidth}
\label{tab:CNN-STF}
\end{table}

Although these CNN-based STF approaches demonstrate strong performance, none of them were originally designed or validated for LST. Applying them directly to LST often leads to model instability or divergence, primarily because its spatial-temporal dynamics. To address these challenges, a few CNN-based STF methods have been specifically developed for LST estimation. Yin et al. \cite{yin2020spatiotemporal} introduced STTFN, a dual-CNN framework designed to accommodate the spatio-temporal variability and higher non-linearity of LST. The model uses two independently trained networks (forwardCNN and backwardCNN) to estimate \( \hat{X}_2(s,t_3) \) and \( \hat{X}_2(s,t_1) \) from \( P_1 \) and \( P_3 \), respectively. At the prediction stage, both networks process \( P_2 \), and their outputs are combined using a temporal weighting function to produce the final LST estimate at \( t_2 \). The Huber loss is adopted to reduce the influence of extreme temperature values. More recently, Sun et al.~\cite{sun2023supervised} proposed a cascade STF architecture that explicitly leverages multiple temporal pairs to better capture the temporal smoothness of LST. The framework consists of a supervised module generating initial spatially consistent predictions, followed by a self-supervised refinement block that constrains temporal behavior at prediction dates. In addition to content loss, the method introduces cycle-consistency and temporal-consistency losses, which are particularly important for LST due to its stronger temporal variation compared to SR.

From Table~\ref{tab:CNN-STF}, several trends emerge for CNN-based STF methods. Most approaches rely on MODIS-Landsat satellite pairs. Many methods, including \cite{song2018spatiotemporal} and \cite{zheng2019spatiotemporal}, use two temporal satellite pairs, whereas others, such as \cite{tan2018deriving} operate with a single satellite pair. Content-based losses (e.g., MSE) are the predominant optimization criteria. A large fraction of studies depend on handcrafted datasets, and only a few, such as \cite{tan2018deriving}, provide open-source implementation, revealing a reproducibility gap. Importantly, only two CNN-based methods~\cite{yin2020spatiotemporal, sun2023supervised} have been validated on LST data.

\subsection{Autoencoder}

AEs are unsupervised neural networks designed to learn compact and informative representations of input data~\cite{rumelhart1986learning, bank2023machine}. They consist of an \textit{encoder}, which maps the input \(X = \{x_i\}_{i=1}^n\)  to a latent representation \(H = \{h_i\}_{i=1}^n\) , and a \textit{decoder}, which reconstructs the input from this latent space. The network is trained to minimize the reconstruction error between the original and reconstructed input as defined in Equation \ref{eq:minAE}.

\begin{equation}
\min _\theta J_{AE}(\theta) = \min _\theta \sum_{i=1}^n l\left(x_i, x_i^{\prime}\right) = \min _\theta \sum_{i=1}^n l\left(x_i, g_\theta\left(f_\theta\left(x_i\right)\right)\right)
\label{eq:minAE}
\end{equation}

\noindent Here, \(x_i\) denotes the \(i\)-th input sample, \(x_i'\) is its reconstruction, and \(n\) is the total number of samples in the training dataset. The encoder function \(f_\theta\) transforms \(x_i\) into a latent representation \(h_i\), while the decoder function \(g_\theta\) reconstructs the input from this latent representation. The parameters \(\theta\) include all weights and biases of both encoder and decoder networks. Stacked AEs (SAEs) extend this architecture by using multiple encoding and decoding layers to capture more complex hierarchical features~\cite{hinton2006fast}. Variants such as Sparse~\cite{ng2011sparse}, Contractive~\cite{rifai2011contractive}, Denoising~\cite{vincent2010stacked}, Variational~\cite{an2015variational}, Convolutional AEs~\cite{semeniuta2017hybrid}, and Recurrent AEs~\cite{nguyen2021forecasting}, have been developed to improve robustness, noise handling, or temporal and spatial modeling.

\begin{figure}[H]
\centering
\includegraphics[width=12cm]{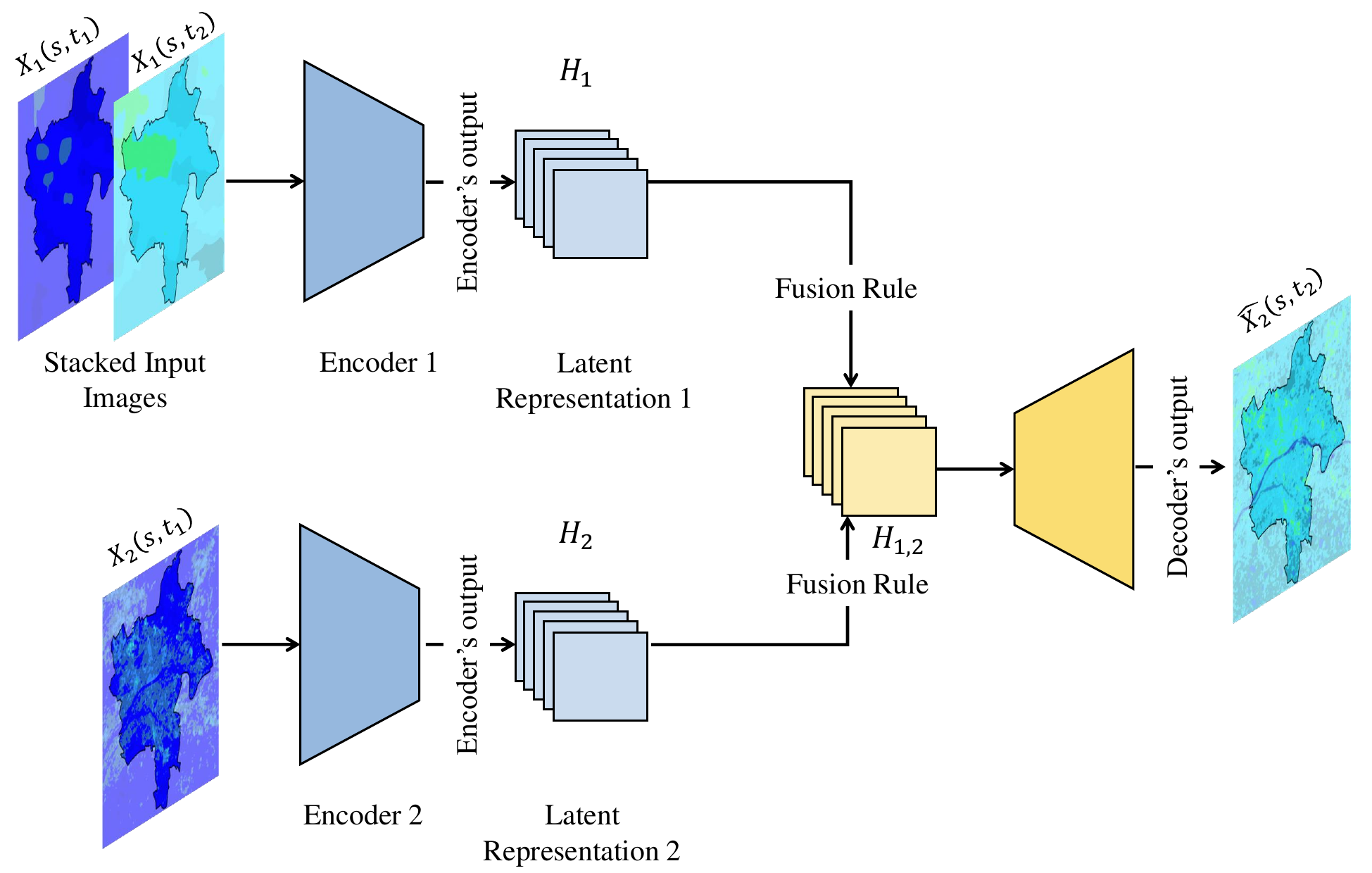}
\caption{Typical architecture of an AE-based STF method using a single pair of LST data $P_1$. It consists of two encoders: a spatial feature encoder and a temporal variation encoder, along with a decoder. The fusion occurs between the latent representations of the outputs from both encoders. The fusion rule can be element-wise addition, multiplication, concatenation, attention, etc. \label{fig:STF-ae}}
\end{figure}

AE-based STF methods extend traditional AEs to model both spatial and temporal relationships between LSHT and HSLT image pairs. Although AEs are primarily unsupervised, STF applications adapt them for supervised tasks using an \textit{encoder-fusion-decoder} design, as illustrated in Figure \ref{fig:STF-ae}. In this architecture, a spatial encoder processes the high-resolution satellite image to produce a latent representation \(H_1\), while a temporal encoder compresses two low-resolution satellite images into \(H_2\). The latent representations \(H_1\) and \(H_2\) are then fused through a tailored mechanism to capture spatio-temporal dependencies. The merged latent representation is decoded to reconstruct the predicted high-resolution satellite image \(\hat{X}_2\), which is compared to the true target \(X_2\) to compute the loss. Table \ref{tab:AE-STF} provides a summary of AE-based DL approaches for STF. Tan et al.~\cite{tan2019enhanced} introduced EDCSTFN, which employs a basic AE architecture and a compound loss combining content, feature, and vision components. 

For LST estimation, Chen et al.~\cite{chen2022spatiotemporal} proposed a Conditional Variational AE (CVAE). The encoder utilizes a multi-kernel convolutional transformer to extract global features, while the decoder reconstructs the fused LST image with a simple convolutional structure. A compound loss integrating variational inference and noise mitigation improves robustness against outliers and ensures high-quality reconstruction. Note that GAN-based AE methods are more appropriately categorized under GAN models and are therefore excluded from this AE section for clarity.

Table~\ref{tab:AE-STF} presents a comparative overview of AE-based STF methods. Most approaches rely on two temporal satellite image pairs~\cite{tan2019enhanced, chen2022spatiotemporal}. Open-source implementations are few, with only \cite{tan2019enhanced} providing accessible code. Validation on LST data remains limited, with \cite{chen2022spatiotemporal} being a notable exception.


\begin{table}[h]
\caption{A comparative overview of AE-based DL methods for STF. For each method, the table lists the satellite sensors used (\textit{Satellites}), whether the method is evaluated on LST data (\textit{LST}), the number of temporal pairs required for training (\textit{Pairs}), the loss functions employed (\textit{Loss Functions}), the evaluation metrics reported (\textit{Evaluation metrics}), the type of datasets used (\textit{Datasets}), and the availability of implementation code (\textit{Code}).}
\begin{adjustwidth}{-1.5cm}{-1.5cm}
\renewcommand{\arraystretch}{1}
\begin{tabularx}{\fulllength}{l@{\hspace{-1pt}}>{\centering\arraybackslash}m{3cm}@{\hspace{-12pt}}C@{\hspace{-30pt}}C@{\hspace{-15pt}}C >{\centering\arraybackslash}m{3.5cm} C@{\hspace{-10pt}}C@{\hspace{-10pt}}}
\toprule
\textbf{Method} & \textbf{Satellites} & \textbf{LST } & \textbf{Pairs} & \textbf{Loss Functions} & \textbf{Evaluation metrics} & \textbf{Datasets} & \textbf{Code} \\
\midrule

~\cite{tan2019enhanced} & MODIS, Landsat & \xmark & 2 & Content, Feature, Vision& RMSE, SAM, ERGAS & Hand-crafted & \href{https://github.com/theonegis/edcstfn}{\cmark} \\

~\cite{chen2022spatiotemporal} & MODIS, FY-4A & \cmark & 2 & Content & RMSE, SSIM, LPIPS & Hand-crafted & \xmark \\

\bottomrule
\end{tabularx}
\end{adjustwidth}
\label{tab:AE-STF}
\end{table}

\subsubsection{Generative Adversarial Networks}

\begin{figure}[h]
\centering
\includegraphics[width=12cm]{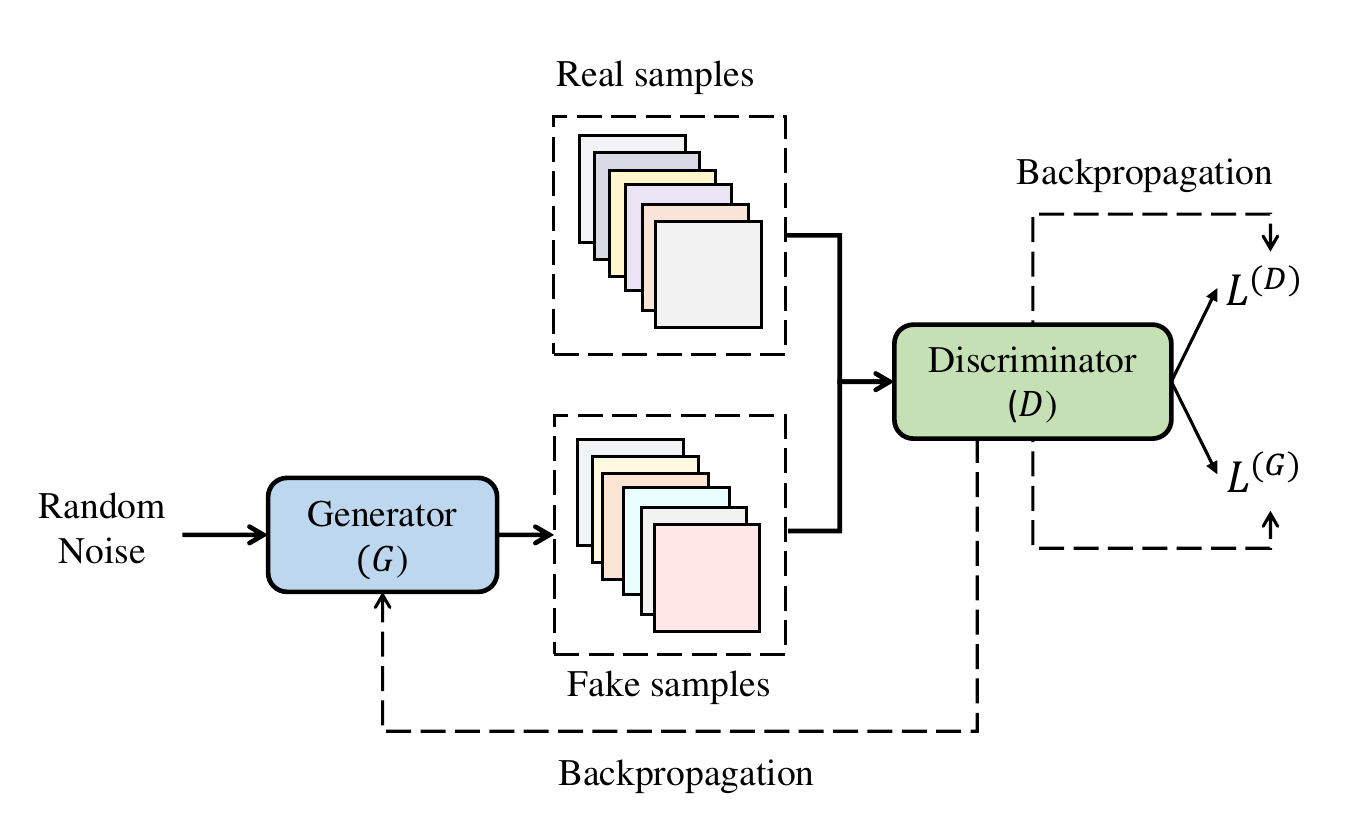}
\caption{Standard GAN architecture. The generator $G$ creates synthetic data from random noise $z$, while the discriminator $D$ evaluates inputs to distinguish real samples $x$ from generated outputs. Both networks are trained adversarially using loss functions $L^{(D)}$ and $L^{(G)}$, with the generator aiming to fool the discriminator and the discriminator attempting to correctly classify real from fake data.\label{fig:GAN}}
\end{figure}

GANs are a class of machine learning models designed for unsupervised learning, where the objective is to produce synthetic data that closely resembles real-world samples.~\cite{goodfellow2014generative}. A GAN consists of two neural networks: a \textit{generator} \(G\), which produces artificial data, and a \textit{discriminator} \(D\), which evaluates whether the data is real or fake~\cite{sharma2024generative, hong2019generative}. The two networks engage in a competitive, adversarial game, in which the generator aims to fool the discriminator, while the discriminator seeks to accurately distinguish real from generated data~\cite{goodfellow2020generative}. This iterative min-max process enables the generator to produce outputs often indistinguishable from real data. GANs have been widely applied to image generation~\cite{zhu2016generative}, super-resolution~\cite{ledig2017photo}, denoising~\cite{zhang2019image}, and image-to-image translation~\cite{zhu2017unpaired}. The basic architecture of a standard GAN is illustrated in Figure~\ref{fig:GAN}.

The discriminator's objective is to maximize the likelihood of assigning high scores to real data and low scores to generated data, as formulated in Equation \ref{eq:lossDF}. Here, \(x \sim \mu(x)\) denotes real samples drawn from the true data distribution, and \(z \sim \gamma(z)\) represents latent noise vectors sampled from a prior distribution used as input for the generator.

\begin{equation}
L^{(D)} = \max_D \left[ \log D(x) + \log\left(1 - D(G(z))\right) \right].
\label{eq:lossDF}
\end{equation}

Conversely, the generator is trained to minimize the discriminator’s ability to distinguish real from generated data, as shown in Equation \ref{eq:lossGF}.

\begin{equation}
L^{(G)} = \min_G \left[ \log D(x) + \log\left(1 - D(G(z))\right) \right]
\label{eq:lossGF}
\end{equation}

Combining these two objectives yields the standard adversarial min-max optimization problem expressed in Equation \ref{eq:minmaxLoss}.

\begin{equation}
L = \min_G \max_D \left[ \log D(x) + \log\left(1 - D(G(z))\right) \right]
\label{eq:minmaxLoss}
\end{equation}

For the full dataset, the objective is expressed using expectations over real data \(x \sim \mu(x)\) and latent variables \(z \sim \gamma(z)\), as presented in Equation \ref{eq:optimproblem2}.

\begin{equation}
\begin{split}
\min_G \max_D V(D,G) &= \min_G \max_D \Big[ \mathbb{E}_{x \sim \mu} [\log D(x)]  
 + \mathbb{E}_{z \sim \gamma} [\log (1 - D(G(z)))] \Big]
\end{split}
\label{eq:optimproblem2}
\end{equation}

GANs have evolved into multiple variants for diverse applications. Fully Connected GANs use simple MLPs for both networks~\cite{goodfellow2014generative}, while Convolutional GAN introduce convolutional layers suitable for image generation~\cite{radford2015unsupervised, denton2015deep}. Conditional GANs (cGANs) allow controlled generation by conditioning on auxiliary information~\cite{mirza2014conditional, chen2016infogan}. GANs with inference models, such as ALI~\cite{dumoulin2016adversarially} and BiGAN~\cite{donahue2016adversarial}, map observed data to latent representations. Adversarial AEs (AAEs) combine AE reconstruction with adversarial learning to improve generative modeling~\cite{kingma2013auto, mescheder2017adversarial}.

\begin{figure}[H]
\centering
\includegraphics[width=12cm]{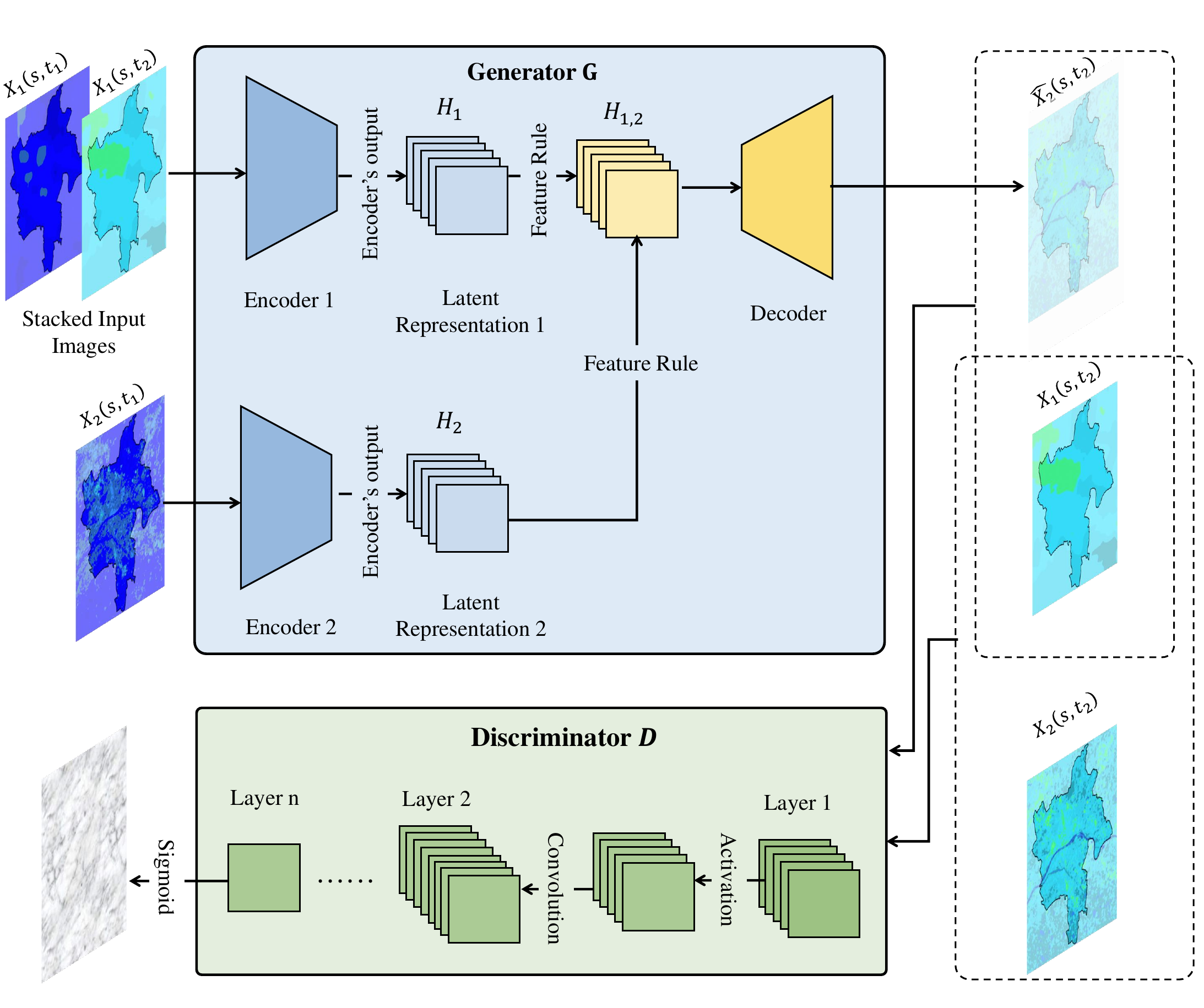}
\caption{Typical architecture of a cGAN-based STF method. It consists of a generator that creates high-resolution fused satellite images using input pairs and a discriminator that distinguishes between real and generated fine-resolution images. The generator includes feature extraction, fusion, and image reconstruction stages, while the discriminator evaluates the fused satellite image consistency. \label{fig:STF-GAN}}
\end{figure}

GAN-based STF models consist of two primary components, as illustrated in Figure~\ref{fig:STF-GAN}: the \textit{generator} and the \textit{discriminator}. The generator is responsible for producing high-resolution fused satellite images using either a single or two temporal satellite pairs ($P_1$ and $P_3$). The discriminator evaluates these outputs by distinguishing real high-resolution satellite images from generated ones. cGANs are the most widely adopted variant for STF, as conditioning the generation process on low-resolution inputs helps maintain spatial and temporal consistency. The generator typically follows a three-stage design:  \textit{feature extraction},  \textit{feature fusion}, and \textit{image reconstruction}. An encoder-decoder architecture is commonly used to enhance spatial resolution during the extraction and reconstruction stages, while the fusion module integrates multi-spatial and multi-temporal features to produce temporally coherent predictions. The discriminator receives as input both the coarse-resolution satellite image at the target date and either the real or the generated fine-resolution satellite image. Its objective is to output a probability indicating whether the pair is real (ground truth) or fused (generated), usually achieved with a sigmoid activation in the final layer. During training, when the discriminator is provided with real pairs (true fine-resolution and coarse-resolution LST images), it is encouraged to classify them as real (label 1). Conversely, when presented with fused pairs (generated fine-resolution and coarse-resolution LST images), it learns to classify them as fake (label 0). Table \ref{tab:GAN-STF} presents a comparative summary of cGAN-based DL techniques for STF. In \cite{chen2020cyclegan}, a CycleGAN architecture was used to model temporal transitions between HSLT satellite images at $k\!-\!1$ and $k\!+\!1$, with final refinements performed using wavelet transforms. A two-stage GAN framework was later proposed in \cite{zhang2020remote}, where the generator relies on residual blocks for high-frequency feature extraction and a feature-level fusion strategy, validated on three MODIS-Landsat datasets (CIA, LGC~\cite{emelyanova2013assessing}, and Shenzhen). Ma et al. \cite{ma2021explicit} modeled spatial, sensor, and temporal inconsistencies using cascaded dual regression networks, four-layer CNNs, and a GAN module, respectively. Several studies have also focused on simplifying the inputs required for GAN-based STF. For example, \cite{tan2021flexible} introduced a cGAN-based model that performs fusion using only two images: a coarse-resolution observation from the prediction date and a fine-resolution reference satellite image from any prior date. Additional improvements include the integration of advanced segmentation and linear injection strategies \cite{zhang2021object}, multilevel feature fusion with attention modules and adaptive instance normalization (MLFF-GAN) \cite{song2022mlff}, as well as robust attention mechanisms designed to handle noisy inputs \cite{tan2022robust}. More recent advances include adaptive multiscale pyramidal architectures with deformable convolutions \cite{pan2023adaptive} and diffusion-based generative models that iteratively refine noisy inputs using dual-stream U-Net architectures \cite{huang2024stfdiff}. These approaches have been evaluated on a diverse set of benchmark datasets, including CIA, LGC~\cite{emelyanova2013assessing}, Tianjin~\cite{li2020spatio}, and E-Smile dataset~\cite{xia2024sossf}. 

For LST estimation, Chen et al.~\cite{chen2025cgmfn} proposed a cGAN-based STF architecture specifically designed for LST generation. The approach constructs an unsupervised generative network, which iteratively generates fine spatio-temporal resolution LST data from reference LST images containing missing values.


Table \ref{tab:GAN-STF} provides a summary of GAN-based STF methods. Most approaches rely on two temporal pairs for predictions, though some, such as \cite{ma2021explicit}, \cite{zhang2021object}, and ~\cite{chen2025cgmfn}, use a single pair. Loss functions typically combine adversarial losses with additional terms, such as content, spectral, or perceptual losses, to improve reconstruction quality and spatial fidelity. Open-source implementations are relatively common, with \cite{tan2021flexible}, \cite{song2022mlff}, and \cite{pan2023adaptive} providing accessible code, which supports reproducibility and further development. Importantly, only one GAN-based STF method~\cite{chen2025cgmfn} has been validated on LST data.

\begin{table}[h]
\caption{A comparative overview of GAN-based DL methods for STF. For each method, the table lists the satellite sensors used (\textit{Satellites}), whether the method is evaluated on LST data (\textit{LST}), the number of temporal pairs required for training (\textit{Pairs}), the loss functions employed (\textit{Loss Functions}), the evaluation metrics reported (\textit{Evaluation metrics}), the type of datasets used (\textit{Datasets}), and the availability of implementation code (\textit{Code}).}
\begin{adjustwidth}{-1.5cm}{-1.5cm}
\renewcommand{\arraystretch}{1}
\begin{tabularx}{\fulllength}{l@{\hspace{-1pt}}>{\centering\arraybackslash}m{3cm}@{\hspace{-12pt}}C@{\hspace{-30pt}}C@{\hspace{-15pt}}C >{\centering\arraybackslash}m{3.5cm} C@{\hspace{-10pt}}C@{\hspace{-10pt}}}
\toprule
\textbf{Method} & \textbf{Satellites} & \textbf{LST } & \textbf{Pairs} & \textbf{Loss Functions} & \textbf{Evaluation metrics} & \textbf{Datasets} & \textbf{Code} \\
\midrule

~\cite{chen2020cyclegan} & MODIS, Landsat & \xmark & 2 &  Adversarial &  RMSE, CC, SSIM, SAM, ERGAS & Hand-crafted & \xmark \\

~\cite{zhang2020remote} & MODIS, Landsat & \xmark & 2 & Content, Adversarial &MAE, RMSE, SSIM, SAM, ERGAS, time& \begin{tabular}[t]{@{}c@{}} CIA, LGC, \\Shenzhen  \end{tabular} & \xmark \\

~\cite{ma2021explicit} & MODIS, Landsat & \xmark & 1 & Content, Vision, Spectral, Adversarial & RMSE, SSIM, SAME, $Q4$ & CIA, LGC & \xmark \\

~\cite{tan2021flexible} & MODIS, Landsat & \xmark & 1/2 &Content, Vision, Feature, Adversarial& MAE, RMSE, SAM, SSIM & CIA, LGC &  \href{https://github.com/theonegis/ganstfm}{\cmark} \\

~\cite{zhang2021object} & MODIS, Landsat & \xmark & 1 & Adversarial &MSE, SSIM, CC, UIQI, ERGAS, SAM& CIA, LGC & \xmark \\

~\cite{song2022mlff} & MODIS, Landsat  & \xmark  & 1 &Content, Vision, Spectral, Adversarial& MAE, RMSE, SAM, SSIM & CIA, LGC & \href{https://github.com/songbingze/MLFF-GAN}{\cmark} \\

~\cite{tan2022robust} & MODIS, Landsat & \xmark & 2 & Content, Vision, Spectral, Feature  & RMSE, SAM, SSIM, ERGAS & CIA, LGC & \href{https://github.com/theonegis/rsfn}{\cmark} \\

~\cite{pan2023adaptive} & MODIS, Landsat & \xmark & 2  & Content,  Vision, Spectral, Adversarial & RMSE, SSIM, PSNR, CC &\begin{tabular}[t]{@{}c@{}} CIA, LGC, \\Tianjin \end{tabular} & \href{https://github.com/xxsfish/AMS-STF.git}{\cmark} \\

~\cite{huang2024stfdiff} &  MODIS, Landsat & \xmark & 2 & Content &  RMSE, SSIM, CC, SAM, ERGAS  &  \begin{tabular}[t]{@{}c@{}} CIA, LGC \\ E-SMILE \end{tabular} &  \href{https://github.com/prowDIY/STF}{\cmark} \\

~\cite{chen2025cgmfn} & MODIS, FY-4A & \cmark & 1 & Content, Adversarial& RMSE, SSIM, LIPIPS& Hand-crafted & \xmark \\

\bottomrule
\end{tabularx}
\end{adjustwidth}
\label{tab:GAN-STF}
\end{table}

\subsection{Vision Transformers}

\begin{figure}[h]
\centering
\includegraphics[width=12cm]{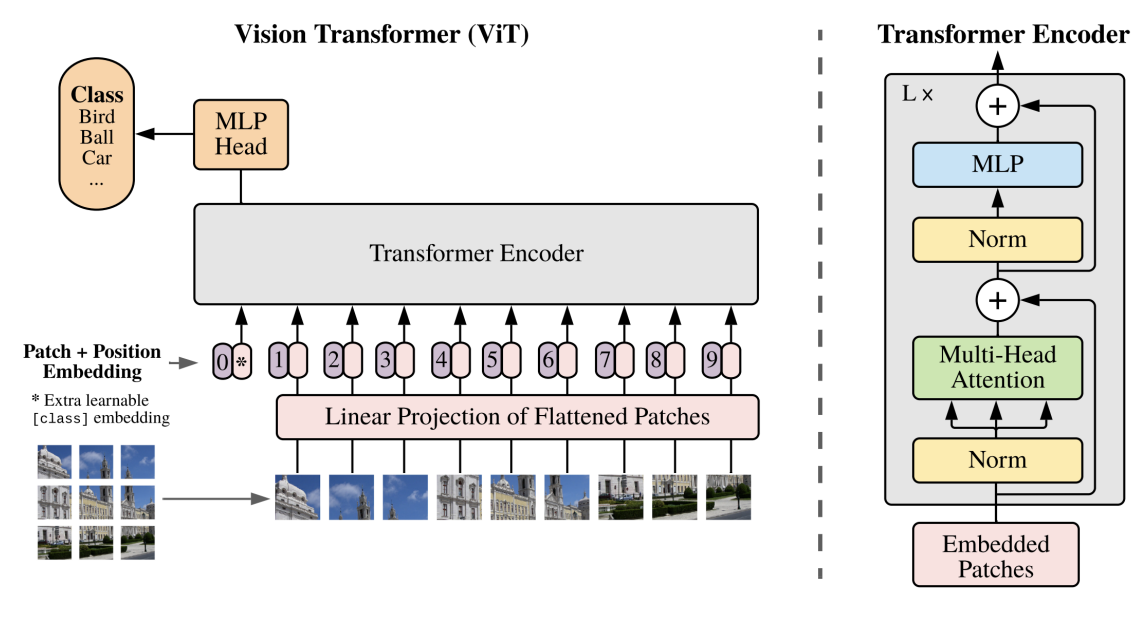}
\caption{Basic architecture of a ViT~\cite{dosovitskiy2020image}. The input image is first divided into fixed-size non-overlapping patches, which are flattened and mapped to patch embeddings through a linear projection. Positional encodings are added to preserve spatial order, and the resulting sequence is fed into a transformer encoder composed of stacked multi-head self-attention and feed-forward layers. The final encoded representation is processed by an MLP head to produce the output.\label{fig:VIT}}
\end{figure}

Transformers were originally introduced for natural language processing to model long-range dependencies through self-attention mechanisms~\cite{vaswani2017attention}, achieving significant success in tasks such as machine translation and text generation~\cite{wolf2020transformers}. Motivated by these results, Dosovitskiy~\cite{dosovitskiy2020image} proposed the ViT, which extended Transformer architectures to computer vision tasks by processing images as sequences of fixed-size patches instead of relying on convolutional operations. Figure~\ref{fig:VIT} illustrates the basic architecture of ViT. In ViT, an input image is divided into non-overlapping patches, each of which is flattened and linearly projected into a patch embedding. These embeddings are combined with positional encodings to preserve spatial information and then passed through a standard Transformer encoder composed of multi-head self-attention and feed-forward networks.

Since the introduction of ViT, numerous architectural variants have been proposed to enhance training efficiency, local feature modeling, and multi-scale representation. DeiT~\cite{touvron2021training} improved data efficiency using knowledge distillation. TNT~\cite{han2021transformer} introduced nested tokens to better capture local structure, while PVT~\cite{wang2021pyramid} incorporated pyramid-style hierarchical features for dense prediction tasks. Swin Transformer~\cite{liu2021swin} addressed the lack of locality in ViT by employing shifted window self-attention. Additional improvements include spatially separable attention in Twins~\cite{chu2021twins}, region-aware token interaction in ViL~\cite{zhang2021multi}, and outlook attention in VOLO~\cite{yuan2022volo}.

\begin{figure}[h]
\begin{adjustwidth}{-1.5cm}{-1.5cm}
\centering
\includegraphics[width=18cm]{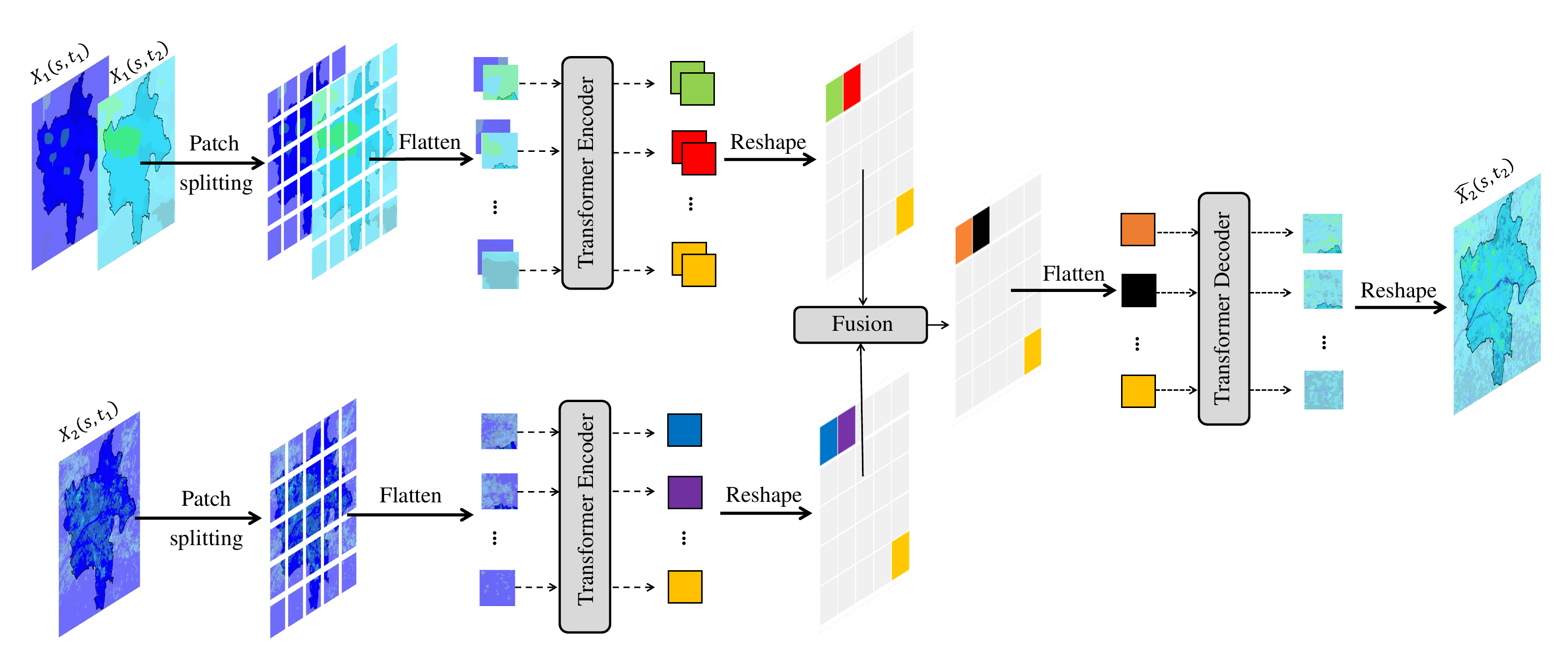} 
\end{adjustwidth}
\caption{Typical architecture of a ViT-based STF method. It consists of a temporal transformer encoder for low-resolution inputs, a spatial feature transformer encoder for high-resolution input, and a transformer decoder for generating high-resolution output. The method processes LST images as patch sequences with positional embeddings. \label{fig:STF-Transformers}}
\end{figure}

ViT-based STF models leverage self-attention mechanisms to capture long-range spatial and temporal dependencies more effectively than convolutional architectures. As shown in Figure~\ref{fig:STF-Transformers}, each input satellite image is first decomposed into a sequence of fixed-size patches, to which positional embeddings are added in order to preserve spatial structure. A spatial transformer encoder processes the high-resolution observation, while a temporal transformer encoder compresses the two coarse-resolution satellite images into latent patch representations. The resulting spatial and temporal sequences are then fused and fed into a transformer decoder, which reconstructs the output patch sequence. After reshaping, the model produces the final high-resolution prediction at the target date. Table~\ref{tab:ViT-STF} provides an overview of ViT-based DL approaches for STF. In \cite{li2021msnet}, a multi-stream architecture combining ViT with CNNs was introduced to jointly exploit global temporal dependencies and local feature extraction. The method was evaluated on the LGC, CIA~\cite{emelyanova2013assessing}, and AHB~\cite{li2020spatio} datasets. In \cite{chen2022swinstfm}, the authors proposed an STF framework based on the swin transformer and integrated linear spectral mixing theory. Li et al.~\cite{li2022enhanced} improved ViT-based STF by introducing an enhanced transformer encoder and dilated convolutions to enlarge the receptive field. MSFusion~\cite{yang2022msfusion} employs a texture ViT to better model spatial structural details, while Jiang and Shao \cite{jiang2024cnn} developed an AE with a multi-kernel convolutional ViT encoder to extract global features across scales. More recently, Benzenati et al.  \cite{benzenati2024stf} introduced STF-Trans, a transformer-based fusion method that requires only a single high-resolution observation at any arbitrary date alongside coarse-resolution temporal inputs, using an AE ViT design to effectively capture long-range dependencies. In addition, Ma et al.~\cite{ma2025sft} proposed SFT-GAN, which integrates a sparse fast ViT within a GAN-based framework to enhance multi-scale feature extraction and spectral consistency. 

For ViT-based STF for LST-specific applications, Hu et al.~\cite{hu2025two} proposed THSTNet, which employs a two-stage Swin ViT architecture with spatiotemporal mapping and a texture converter module to improve the reconstruction of fine-resolution LST.

Table~\ref{tab:ViT-STF} summarizes recent ViT-based methods for STF. Most approaches emerged after 2021, following the introduction of ViT in 2020. Loss functions largely prioritize content preservation, while some methods incorporate additional vision-guided constraints to enhance spatial or spectral fidelity. Several works, such as \cite{chen2022swinstfm} and \cite{ma2025sft}, provide open-source implementations that support reproducibility and further development. Notably, only one ViT-based STF model has been specifically designed and validated for LST data~\cite{hu2025two}.

\begin{table}[h]
\caption{A comparative overview of ViT-based DL methods for STF. For each method, the table lists the satellite sensors used (\textit{Satellites}), whether the method is evaluated on LST data (\textit{LST}), the number of temporal pairs required for training (\textit{Pairs}), the loss functions employed (\textit{Loss Functions}), the evaluation metrics reported (\textit{Evaluation metrics}), the type of datasets used (\textit{Datasets}), and the availability of implementation code (\textit{Code}).}
\begin{adjustwidth}{-1.5cm}{-1.5cm}
\renewcommand{\arraystretch}{1}
\begin{tabularx}{\fulllength}{l@{\hspace{-1pt}}>{\centering\arraybackslash}m{3cm}@{\hspace{-12pt}}C@{\hspace{-30pt}}C@{\hspace{-15pt}}C >{\centering\arraybackslash}m{3.5cm} C@{\hspace{-10pt}}C@{\hspace{-10pt}}}
\toprule
\textbf{Method} & \textbf{Satellites} & \textbf{LST} & \textbf{Pairs} & \textbf{Loss Functions} & \textbf{Evaluation metrics} & \textbf{Datasets} & \textbf{Code} \\
\midrule

~\cite{li2021msnet} & MODIS, Landsat & \xmark & 2 & Content &   RMSE, MSE, CC, SAM, SSIM, ERGAS, PSNR  & CIA, LGC, AHB   & \xmark \\

~\cite{yang2022msfusion} & MODIS, Landsat & \xmark & 2 & Feature &  RMSE, SSIM, ERGAS, SAM, CC  & CIA-LGC, DX   & \xmark \\

~\cite{chen2022swinstfm} &  MODIS, Landsat & \xmark & 1 & Content, Vision  &   RMSE, CC, SAM, SSIM, UIQI  & CIA-LGC &  \href{https://github.com/LouisChen0104/swinstfm.git}{\cmark}\\

~\cite{li2022enhanced} & MODIS, Landsat & \xmark & 1 & Content &   RMSE, MSE, CC, SAM, SSIM, ERGAS, PSNR  & CIA-LGC, AHB   & \xmark\\

~\cite{jiang2024cnn} & MODIS, Landsat & \xmark & 2& Content & MAE, SAM, SSIM, PSNR & CIA, DX & \xmark \\

~\cite{benzenati2024stf} & Planetscope, Pléiades & \xmark & 1/2 & Content, Vision &  RMSE, CC, SAM, SSIM, UIQI  & Hand-crafted &  \xmark\\

~\cite{ma2025sft} & MODIS, Landsat & \xmark & 1 & Content, Vision, Spectral, Adversarial& RMSE, PSNR, ERGAS, SAM, SSIM, UIQI & CIA, LGC, AHB, Tianjin & \href{https://github.com/MaZhaoX/SFT-GAN.git}{\cmark} \\

~\cite{hu2025two} & MODIS, Landsat & \cmark & 1 & Content, Vision& MAE, RMSE, PSNR, $R^2$&Hand-crafted& \href{https://github.com/HuPengHua2021/THSTNet.git}{\cmark}\\

\bottomrule
\end{tabularx}
\end{adjustwidth}
\label{tab:ViT-STF}
\end{table}

\subsection{Recurrent Neural Networks}
RNNs are a class of supervised machine learning models designed to process sequential or time-series data~\cite{rumelhart1986learning}. They incorporate feedback connections that allow information from previous time steps to influence the current output, enabling the network to retain temporal dependencies~\cite{medsker2001recurrent, das2023recurrent}. Variants such as Long Short-Term Memory networks (LSTMs)~\cite{hochreiter1997long}, Bidirectional LSTMs~\cite{graves2005framewise}, Stacked LSTMs~\cite{karevan2018spatio}, and Gated Recurrent Units (GRUs)~\cite{cho2014learning} have been developed to overcome limitations including vanishing gradients, capture long-term dependencies, and reduce computational complexity.

In RNN-based STF, the aim is to leverage temporal dependencies between coarse and fine-resolution satellite images. Given a sequence of $n$ temporal satellite pairs, $P_i$, the RNN is trained to map coarse-resolution inputs to fine-resolution predictions at each time step $t_i$, denoted as $\hat{X}_2(s,t_i)$, for $i \in [1,n]$. After training, the model generates fine-resolution estimates for new time steps, which are then fused with existing coarse-resolution data to reconstruct high-quality satellite image outputs. Table \ref{tab:RNN-STF} gives an overview of RNN-based DL frameworks applied to STF. Yang et al.~\cite{yang2021robust} proposed a hybrid model combining a Super-Resolution CNN with an LSTM. The CNN enhances spatial resolution, while the LSTM captures temporal patterns, particularly for rapid phenological changes. Zhan et al.~\cite{zhan2024time} extended this approach by integrating a UNet for spatial mapping between MODIS and Sentinel-2 images, combined with an LSTM to exploit temporal dynamics for generating high-resolution NDVI data during critical crop growth periods.

Table~\ref{tab:RNN-STF} summarizes RNN-based STF methods. Overall, RNNs are less commonly used in STF compared to other architectures, primarily because STF benefits from models that jointly capture both spatial and temporal dependencies, whereas RNNs are optimized for sequential data. Notably, RNN-based approaches have not yet been explored for LST-specific STF, which highlights a potential avenue for future research contributions.

\begin{table}[h]
\caption{A comparative overview of RNN-based DL methods for STF. For each method, the table lists the satellite sensors used (\textit{Satellites}), whether the method is evaluated on LST data (\textit{LST}), the number of temporal pairs required for training (\textit{Pairs}), the loss functions employed (\textit{Loss Functions}), the evaluation metrics reported (\textit{Evaluation metrics}), the type of datasets used (\textit{Datasets}), and the availability of implementation code (\textit{Code}).}
\begin{adjustwidth}{-1.5cm}{-1.5cm}
\renewcommand{\arraystretch}{1}
\begin{tabularx}{\fulllength}{l@{\hspace{-1pt}}>{\centering\arraybackslash}m{3cm}@{\hspace{-12pt}}C@{\hspace{-30pt}}C@{\hspace{-15pt}}C >{\centering\arraybackslash}m{3.5cm} C@{\hspace{-10pt}}C@{\hspace{-10pt}}}
\toprule
\textbf{Method} & \textbf{Satellites} & \textbf{LST} & \textbf{Pairs} & \textbf{Loss Functions} & \textbf{Evaluation metrics} & \textbf{Datasets} & \textbf{Code} \\
\midrule

~\cite{yang2021robust} & MODIS, Landsat & \xmark & 2 &  Content & RMSE, ERGAS, SAM & Hand-crafted & \xmark \\

~\cite{meng2022spatio} & MODIS, Landsat & \xmark & 2 &  Content, Vision & RMSE, SAM, SSIM & CIA, LGC & \xmark \\

~\cite{zhan2024time} & MODIS, Sentinel & \xmark & n & Content & RMSE, SSIM& Hand-crafted  & \xmark \\

\bottomrule
\end{tabularx}
\end{adjustwidth}
\label{tab:RNN-STF}
\end{table}

\subsection{Learning Paradigm}
Various learning paradigms have been applied to STF, and they can be grouped into four main categories: \textit{supervised}, \textit{unsupervised}, \textit{self-supervised}, and \textit{collaborative learning}. These categories differ in how they leverage available high-resolution data.

\begin{enumerate}
    \item \textit{Supervised learning.}  
     This paradigm relies on paired training samples, where fine-resolution observations ${X}_2(s, t_2)$ are available as targets. Most existing STF models fall into this category. While supervised learning has proven effective, its dependence on cloud-free, fine-resolution LST data limits scalability in real LST applications.

    \item \textit{Unsupervised learning.}  
    Here, the model is trained without fine-resolution labels, meaning ${X}_2(s, t_2)$ is unknown. Only one recent study has explored an unsupervised STF formulation~\cite{yu2024unsupervised}. This direction is especially promising for LST.

    \item \textit{Self-supervised learning.}  
    Positioned between supervised and unsupervised paradigms, self-supervised learning creates proxy tasks or pseudo-labels directly from the data~\cite{rani2023self}. To date, only one STF method has adopted this strategy~\cite{sun2023supervised}. Extending self-supervised schemes to LST STF remains largely unexplored and could help reduce reliance on scarce fine-resolution LST images.

    \item \textit{Collaborative learning.}  
    This strategy treats STF as a cooperative process, where different learners interact to improve fusion quality~\cite{laal2012benefits}. Only one study has explicitly framed STF in this way~\cite{meng2022spatio}. Such paradigms could be beneficial for LST, as they may better exploit complementary cues between coarse and fine thermal observations.

\end{enumerate}

\subsection{Training Strategy}
Training strategies refer to the techniques employed to enhance the performance and stability of neural networks during optimization. In STF, these strategies play a key role in capturing spatial-temporal dependencies, especially for LST, where models must handle strong thermal spatio-temporal variability. Existing STF methods generally adopt four main training strategies: \textit{residual learning}, \textit{attention mechanisms}, \textit{normalization}, and \textit{dropout}.

\begin{enumerate}

    \item \textit{Residual learning.}    Residual learning~\cite{he2016deep} introduces skip connections that allow the network to learn a residual function instead of the full mapping, which stabilizes the optimization of DL architectures~\cite{shafiq2022deep}.  
    The residual formulation is defined in Equation~\ref{eq:residual1}.
    \begin{equation}
        F(x) := H(x) - x
        \label{eq:residual1}
    \end{equation}

     where \( H(x) \) is the desired underlying function, and \( F(x) \) is the residual mapping. The output of the residual block becomes \(F(x) + x = H(x)\). As shown in Table~\ref{tab:training_strategy}, residual learning is the most common strategy across STF methods due to its ability to preserve essential spatial and temporal structures in LST data.

    \item \textit{Attention mechanisms.}  
    Attention mechanisms enable a model to focus on the most informative components of the input. In STF, four forms are used: \textit{channel}, \textit{spatial}, \textit{temporal}, and \textit{feature} attention. Channel attention assigns importance to individual spectral bands, although its usefulness is limited for LST-focused STF because LST data generally contains only one thermal band. Spatial attention highlights salient spatial regions, helping the model detect areas with strong temperature variability or sharp thermal gradients. Temporal attention emphasizes key time steps, allowing the network to capture rapid LST fluctuations and short-term dynamics. Feature attention evaluates the relevance of entire feature maps. As summarized in Table~\ref{tab:training_strategy}, all ViT-based STF methods incorporate spatial attention, consistent with the fundamental role of attention in ViT architectures.
    
    \item \textit{Normalization.}  
    Normalization refers to a set of transformations applied to stabilize and accelerate model training by enforcing desired statistical properties such as centering, scaling, or decorrelation~\cite{huang2023normalization}. In STF, five main normalization strategies are encountered. \textit{Batch Normalization (BN)}~\cite{ioffe2015batch} mitigates internal covariate shift by standardizing activations within each mini-batch. Thus, given an activation \( a \), BN computes its normalized form as shown in Equation~\ref{eq:batch_norm}.  
    
    \begin{equation}
        \hat{a}^{(i)} = \frac{a^{(i)} - \mu}{\sqrt{\sigma^2 + \epsilon}}
        \label{eq:batch_norm}
    \end{equation}

    where \( \mu \) and \( \sigma^2 \) denote the mini-batch mean and variance, and \( \epsilon > 0 \) ensures numerical stability. Group Normalization (GN)~\cite{wu2018group} standardizes activations within predefined groups. Instance Normalization (IN)~\cite{ulyanov2016instance} normalizes each sample independently and is used to reduce contrast-related variations. Spectral Normalization (SN)~\cite{miyato2018spectral}, mainly applied in GAN-based STF methods. It stabilizes the discriminator training by constraining the Lipschitz constant of weight matrices. Finally, Switchable Normalization (SwN)~\cite{luo2018differentiable} combines three types of statistics: channel-wise, layer-wise, and mini-batch-wise. As shown in Table~\ref{tab:training_strategy}, STF methods vary widely in their choice of normalization strategy, reflecting different architectural needs and constraints, especially when dealing with LST data.
    
    \item \textit{Dropout.}  
    Dropout randomly deactivates neurons during training to reduce overfitting~\cite{salehin2023review}. Although only a few STF approaches use dropout (Table~\ref{tab:training_strategy}), it remains a promising direction for LST STF, where limited high-resolution observations can make models prone to overfitting.

\end{enumerate}


\begin{table}[h]
\centering
\caption{Overview of training strategies employed in state-of-the-art STF methods.The table categorizes strategies into residual learning, attention mechanisms, normalization techniques, and dropout, and lists the corresponding methods that utilize each strategy. For attention and normalization, subtypes are also indicated to highlight specific implementations across different STF approaches.}
\renewcommand{\arraystretch}{1} 
\begin{tabular}{ll}
\toprule
\textbf{Training Strategy} & \textbf{List of methods} \\

\midrule
\textbf{Residual Learning} &\begin{tabular}[t]{@{}l@{}} 
~\cite{song2018spatiotemporal},~\cite{zheng2019spatiotemporal},~\cite{yin2020spatiotemporal},~\cite{wang2020spatiotemporal},~\cite{li2020new},~\cite{sun2023supervised},~\cite{tan2019enhanced},  
~\cite{chen2022spatiotemporal}, ~\cite{zhang2020remote},~\cite{ma2021explicit},~\cite{tan2021flexible},~\cite{song2022mlff},\\~\cite{tan2022robust},
~\cite{pan2023adaptive},~\cite{huang2024stfdiff}, ~\cite{li2021msnet},
~\cite{li2022enhanced},~\cite{jiang2024cnn}~\cite{hu2025two} 
\end{tabular}
\\
\midrule

\textbf{Attention Mechanism } & \begin{tabular}[t]{@{}l@{}}

\textbf{Channel Attention: } ~\cite{qin2022mustfn},~\cite{ma2021explicit},~\cite{jiang2024cnn},~\cite{ma2025sft}  \\

\textbf{Spatial Attention: } ~\cite{huang2024stfdiff},~\cite{li2021msnet},~\cite{yang2022msfusion},~\cite{chen2022swinstfm},
~\cite{li2022enhanced},~\cite{jiang2024cnn},  ~\cite{benzenati2024stf},~\cite{hu2025two} \\

\textbf{Temporal Attention: } ~\cite{zhang2021object},~\cite{song2022mlff},  ~\cite{tan2022robust} \\

\textbf{Feature Attention: } ~\cite{song2022mlff}

\end{tabular}
\\
\midrule

\textbf{Normalization } & \begin{tabular}[t]{@{}l@{}} 

\textbf{Batch Normalization: } ~\cite{song2018spatiotemporal},~\cite{yin2020spatiotemporal},~\cite{qin2022mustfn}, 
~\cite{chen2022spatiotemporal},~\cite{zhang2020remote},~\cite{ma2021explicit},~\cite{tan2021flexible},\\~\cite{tan2022robust},~\cite{pan2023adaptive},~\cite{zhan2024time} \\

\textbf{Group Normalization: } ~\cite{huang2024stfdiff},~\cite{jiang2024cnn} \\

\textbf{Instance Normalization: } ~\cite{song2022mlff} \\

\textbf{Spectral Normalization: } ~\cite{ma2021explicit},~\cite{tan2021flexible},~\cite{tan2022robust},~\cite{chen2025cgmfn}  \\

\textbf{Switchable Normalization: }~\cite{ma2021explicit},~\cite{tan2021flexible},~\cite{yang2022msfusion}

\end{tabular}
\\
\midrule
\textbf{Dropout} & ~\cite{pan2023adaptive},~\cite{li2021msnet},~\cite{li2022enhanced},~\cite{yang2021robust},~\cite{zhan2024time},~\cite{chen2025cgmfn}  \\

\bottomrule
\end{tabular}
\label{tab:training_strategy}
\end{table}

\subsection{Incorporation of Pre-trained Models}

Incorporating pre-trained models is a widely used strategy in DL to enhance model generalization, improve convergence, and mitigate overfitting~\cite{ying2019overview, chen2021pre}. This is particularly relevant in STF, including LST-specific applications, where paired high and low-resolution datasets are limited. Pre-training can be applied at two main levels (Table \ref{tab:pretraining}):

\begin{enumerate}
    \item \textit{Feature Extraction:} In this strategy, pre-trained models are used solely to extract informative features from the input data without further fine-tuning~\cite{puls2023evaluation}. Within STF, feature extraction is often employed to compute spectral or perceptual losses (see Sections~\ref{sec:feature} and~\ref{sec:Spectral}). 

    \item \textit{Transfer Learning:} Transfer learning aims to improve performance on a target task by adapting knowledge from a related source domain~\cite{weiss2016survey}. Formally, given a source dataset \( \mathcal{D}_S \) with task \( \mathcal{T}_S \) and a target dataset \( \mathcal{D}_T \) with task \( \mathcal{T}_T \), transfer learning seeks to enhance the target predictive function \( f_T(\cdot) \) by utilizing knowledge from \( \mathcal{D}_S \) and \( \mathcal{T}_S \), where \( \mathcal{D}_S \neq \mathcal{D}_T \) or \( \mathcal{T}_S \neq \mathcal{T}_T \).
    In STF for LST, Chen et al.~\cite{chen2022spatiotemporal} pre-trained an autoencoder on simulated LST data generated by downscaling MODIS measurements to 4 km resolution via pixel aggregation, and transferred the learned parameters to initialize their fusion framework. Additionally, Huang et al.~\cite{huang2024stfdiff} proposed a fine-tuning strategy to adapt pre-trained models to new regions, but did not employ pre-training for initial model training, and thus is not included under this category.
\end{enumerate}

\begin{table}[h]
\centering
\caption{Overview of pre-training strategies adopted in state-of-the-art STF methods. The table categorizes approaches into feature extraction and transfer learning. Corresponding methods employing each strategy are listed.} 
\renewcommand{\arraystretch}{1} 
\begin{tabular}{ll}
\toprule
\textbf{Training Strategy} & \textbf{List of methods} \\

\midrule
\textbf{Feature Extraction} &~\cite{tan2019enhanced}, ~\cite{ma2021explicit}, ~\cite{tan2021flexible}, ~\cite{song2022mlff}, ~\cite{tan2022robust}, ~\cite{pan2023adaptive}, ~\cite{yang2022msfusion}
\\
\midrule

\textbf{Transfer Learning} &~\cite{chen2022spatiotemporal}
\\

\bottomrule
\end{tabular}
\label{tab:pretraining}
\end{table}

\section{Experiment Analysis and Results}
\label{sec:test}
This section presents the experimental analysis and results. We begin by defining the ROI and providing a comprehensive description of our LST dataset (STF-LST). Next, we compare representative state-of-the-art STF approaches both quantitatively and qualitatively, and we highlight the challenges posed by the domain shift from SR to LST data.

\subsection{Region of Interest}

\begin{figure}[H]
\centering
\includegraphics[width=13cm]{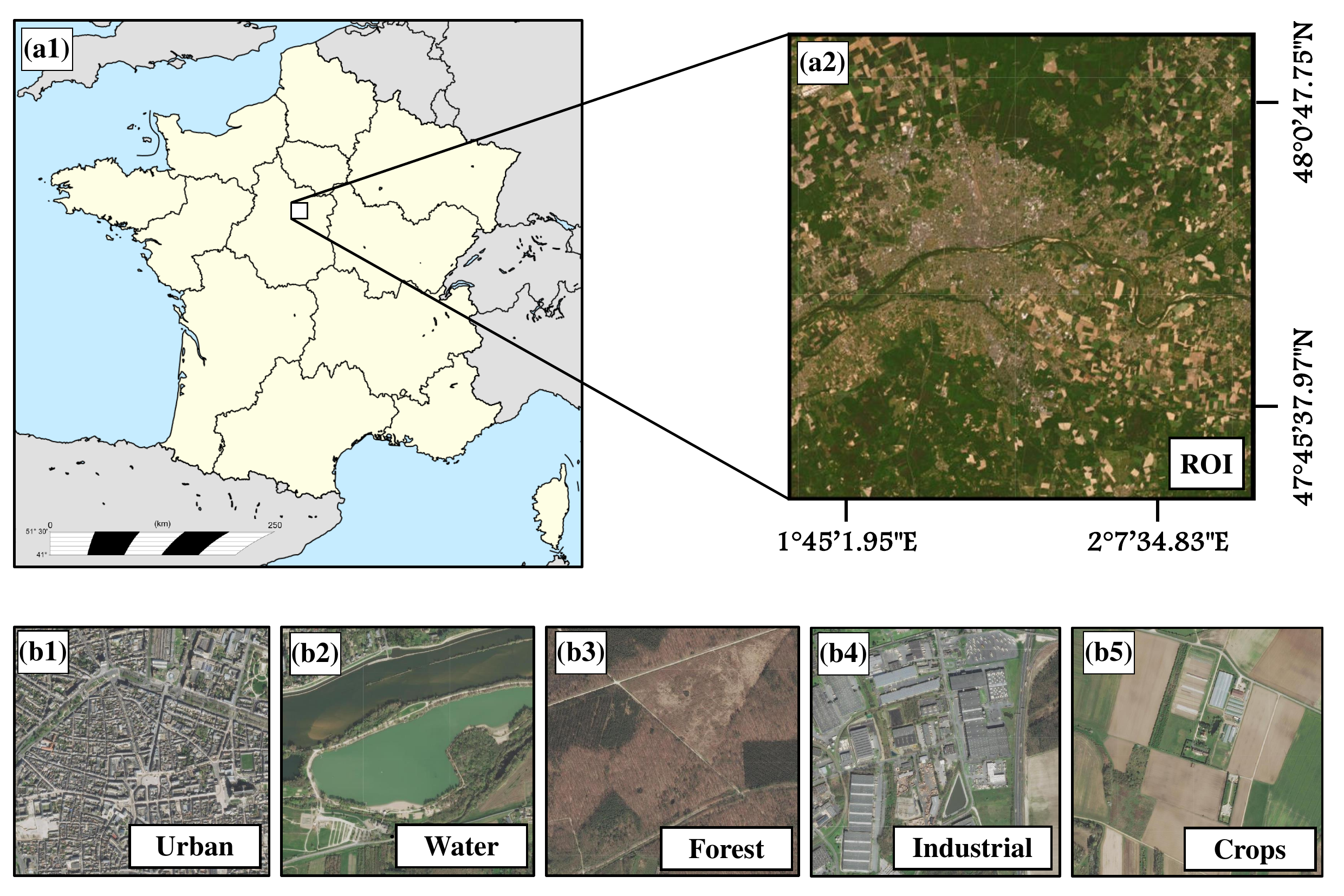}
\caption{Geographic overview of the ROI. (a1) Position of the ROI within France. (a2) High-resolution satellite image of the ROI. (b1)-(b5) Representative examples of key land cover categories observed in the ROI: urban, water, forest, industrial, and agricultural zones.\label{fig:ROI}}
\end{figure}

The ROI is located within Orléans Métropole in the Centre-Val de Loire region of France. As shown in Figure~\ref{fig:ROI}(a1), it extends from longitudes 1.7505°E to 2.1263°E and latitudes 47.7605°N to 48.0133°N, covering approximately 334\,km$^2$. The Loire River, France’s longest watercourse, crosses the ROI and significantly modulates its local microclimate, introducing marked spatial variations in LST (Figure~\ref{fig:ROI}(a2)). The ROI exhibits a high degree of land cover heterogeneity, including dense urban cores, water bodies, forested surfaces, industrial areas, and agricultural croplands(Figure~\ref{fig:ROI}(b1)-(b5)). This mixture produces strong contrasts in radiative properties, emissivity, and thermal inertia. Such conditions are particularly challenging for LST STF, as they generate sharp temperature gradients, nonlinear temporal dynamics, and heterogeneous spatial patterns that must be reconstructed at fine resolution. Orléans Métropole, with nearly 280,000 inhabitants across 22 municipalities, presents a representative mid-sized European urban environment where UHI, river-induced cooling, and peri-urban agricultural influences coexist. This diversity makes the ROI an appropriate benchmark for evaluating the robustness and generalization ability of STF models for LST data.

\subsection{Satellite Data}

This study relies on two complementary satellite products accessed through the Google Earth Engine (GEE) platform~\cite{gorelick2017google}: (i) MODIS/Terra LST and Emissivity Daily Global 1 km (MOD11A1, Collection~6.1) for coarse-resolution observations, and (ii) Landsat~8 USGS Level-2 Collection~2 Tier~1 for fine-resolution LST. These datasets were selected for two main reasons. First, both satellites acquire data during mid-morning overpasses, which ensures that MODIS and Landsat capture surface and atmospheric conditions that are highly comparable in illumination, temperature, and emissivity. Second, when combined, they provide complementary strengths: MODIS offers daily revisit frequency at 1~km resolution, whereas Landsat~8 provides detailed 30~m spatial information but with a 16-day revisit cycle. Leveraging both of them provides fine-resolution daily LST. MODIS LST values were extracted from the \textit{LST\_Day\_1km} band derived using the split window algorithm, which yields an RMSE below 2~$^\circ$C across most land cover types~\cite{duan2019validation}. Landsat~8 LST was derived from the \textit{ST\_B10} thermal band using the single-channel algorithm, with reported accuracy around 1.5~$^\circ$C~\cite{jimenez2014land}. Both products in GEE include standard atmospheric corrections, emissivity adjustments, and quality assurance masks, which reduces preprocessing inconsistencies~\cite{wan1999modis,ermida2020google}.

A total of $51$ MODIS-Landsat paired observations were collected between 14 Apr 2013 and 05 Oct 2024. Only scenes with acceptable cloud levels were retained, <80\% cloud cover for Landsat~8 and <90\% for MODIS. Each pair corresponds to a date on which the MODIS observation and the Landsat~8 acquisition overlapped spatially and temporally over the defined ROI. All $51$ pairs used in the experiments are listed in Table \ref{tab:dataset_samples}. The dataset was divided into train, validation, and test, and was structured into temporal triplets of the form $(t_i, t_{i+1}, t_{i+2})$ . Within the training set, successive triplets were allowed to overlap to increase the density and variability of training instances. For instance, the sample $(t_1, t_2, t_3)$ is followed by $(t_2, t_3, t_4)$. This is widely used in temporal learning settings, as it exposes the model to a richer set of temporal transitions while preserving the physical coherence of the series. However, to avoid temporal leakage and ensure a fair evaluation, no overlap was allowed between the training, validation, and test subsets. The final distribution consists of 34 triplet samples for training, 4 for validation, and 8 for testing.

\begin{table}[t]
\caption{Paired MODIS/Terra and Landsat~8 observations used in this study. Each entry lists the sample number, acquisition date, and overpass times for MODIS and Landsat~8. Only scenes with acceptable cloud coverage were retained (<90\% for MODIS, <80\% for Landsat~8).}
\begin{adjustwidth}{-1.5cm}{-1.5cm}
\renewcommand{\arraystretch}{1}
\begin{tabularx}{\fulllength}{cccc|cccc}
\toprule
\textbf{Sample No.} & \textbf{Date} & \textbf{MODIS/Terra} & \textbf{Landsat8} & \textbf{Sample No.} & \textbf{Date} & \textbf{MODIS/Terra} & \textbf{Landsat8}  \\
\midrule

1 & 14 Apr 2013 & 11:54 & 10:43 & 27 & 21 Oct 2018 & 11:54 & 10:41 \\

2 & 01 Jun 2013 & 11:08 & 10:43  & 28 & 26 Feb 2019 & 11:54 & 10:40\\ 

3 & 04 Aug 2013 & 11:54 & 10:43 & 29 & 02 Jun 2019 & 11:54 & 10:40\\

4 & 20 Aug 2013 & 11:54 & 10:43 & 30 & 04 Jul 2019 & 11:54 & 10:41 \\

5 & 05 Sep 2013 & 11:54 & 10:42 & 31 & 06 Sep 2019 & 11:54 & 10:41  \\

6 & 10 Dec 2013 & 11:54 & 10:42 & 32 & 01 Apr 2020 & 11:54 & 10:40 \\

7 & 16 Mar 2014 & 11:54 & 10:41 & 33 & 19 May 2020 & 11:54 & 10:40\\

8 & 17 Apr 2014 & 11:54 & 10:41 & 34 & 22 Jul 2020 & 11:54 & 10:41 \\

9 & 19 May 2014 & 11:54 & 10:40 & 35 & 07 Aug 2020 & 11:54 & 10:41 \\

10& 08 Sep 2014 & 11:54 & 10:41 & 36 & 27 Nov 2020 & 11:54 & 10:41\\

11& 24 Sep 2014 & 11:54 & 10:41 & 37 & 04 Apr 2021 & 11:54 & 10:40\\

12& 20 Apr 2015 & 11:54 & 10:40 & 38 & 10 Aug 2021 & 11:03 & 10:41\\

13& 10 Aug 2015 & 11:54 & 10:41 & 39 & 06 Mar 2022 & 11:48 & 10:41 \\

14 & 11 Sep 2015 & 11:47 & 10:41 & 40 & 22 Mar 2022 & 11:48 & 10:41 \\

15 & 09 Jun 2016 & 11:54 & 10:40 & 41 & 09 May 2022 & 11:42 & 10:41\\

16 & 12 Aug 2016 & 11:54 & 10:41 & 42 & 13 Aug 2022 & 11:42 & 10:41 \\ 

17 & 15 Oct 2016 & 11:54 & 10:41 & 43 & 29 Aug 2022 & 11:42 & 10:41 \\ 

18 & 31 Oct 2016 & 11:54 & 10:41 & 44 & 30 Sep 2022 & 11:26 & 10:41  \\ 

19 & 19 Jan 2017 & 11:54 & 10:41 & 45 & 01 Nov 2022 & 11:12 & 10:41  \\ 

20 & 09 Apr 2017 & 11:54 & 10:40 & 46 & 28 May 2023 & 11:10 & 10:40 \\ 

21 & 12 Jun 2017 & 11:54 & 10:40 &47 & 13 Jun 2023 & 10:36 & 10:40 \\ 

22 & 23 Feb 2018 & 11:54 & 10:40 & 48 & 19 Oct 2023 & 10:55 & 10:41 \\ 

23 & 11 Mar 2018 & 10:40 & 10:40 & 49 & 12 Apr 2024 & 10:34 & 10:40 \\ 

24 & 02 Aug 2018 & 10:48 & 10:40 & 50 & 19 Sep 2024 & 10:00 & 10:41 \\ 

25 & 18 Aug 2018 & 11:55 & 10:40 & 51 & 05 Oct 2024 & 10:48 & 10:41\\ 

26 & 05 Oct 2018 & 11:53 & 10:41 \\

\bottomrule
\end{tabularx}
\end{adjustwidth}
\label{tab:dataset_samples}
\end{table}

LST gaps were corrected using interpolation techniques. Landsat~8 LST gaps were reconstructed using temporal interpolation over a 32-day window by exploiting two valid observations before and after each missing acquisition to recover smoothly varying temperature patterns. Remaining invalid pixels were filled using an adaptive spatial strategy based on a focal-mean filter, which expanded iteratively until at least one valid neighbor was available. MODIS LST gaps were reconstructed by applying only spatial gap filling using the same adaptive focal-mean approach. After gap correction, all MODIS LST images were resampled to 30~m using bicubic interpolation to ensure pixel-wise alignment with Landsat 8 and to harmonize the spatial dimensions required for STF.

All LST images of size $950 \times 950$ pixels were divided into fixed-size patches. A patch size of $95 \times 95$ with a stride of 20 pixels was selected to balance spatial context against computational cost, ensuring that each patch retained sufficient thermal and textural variability without mixing heterogeneous land-cover patterns excessively. This procedure generated $62866$ training patches and $7396$ validation patches from the available scenes.

\subsection{Quantitative comparison}

The quantitative evaluation relies on six widely used metrics: RMSE, ERGAS, SSIM, PSNR, SAM, and CC, introduced earlier in Section~\ref{sec:evaluation}. Together, they capture complementary aspects of fusion quality, including pixel-wise accuracy, spectral consistency, structural similarity, and global fidelity.  To ensure a representative and balanced comparison, we selected approaches that are established in the STF literature with demonstrated effectiveness and that provide publicly available implementations to guarantee reproducibility. Following these criteria, four methods were retained. ESTARFM~\cite{zhu2010enhanced}, originally designed for SR-based STF, serves as a classical reference for weight-based methods. STTFN~\cite{yin2020spatiotemporal}, a CNN-based architecture explicitly developed for LST STF. EDCSTFN~\cite{tan2019enhanced}, an AE-based SRF framework, enables testing the transferability of SR-oriented STF to LST data. MLFF-GAN~\cite{song2022mlff}, a GAN-based model integrating multi-level feature fusion, represents a modern generative approach with strong performance on SR tasks. All DL-based STF models were trained for 200~epochs with a batch size of~32. The learning rates were set to $1.5 \times 10^{-5}$ for STTFN, $10^{-3}$ for EDCSTFN, and $2 \times 10^{-4}$ for MLFF-GAN. All experiments were conducted on an NVIDIA RTX~A6000 GPU.

Quantitative results obtained from these evaluations are reported in Table~\ref{tab:quantitive}. On 29~Aug~2022, ESTARFM yields the best scores across all metrics among the compared methods. Nevertheless, its RMSE remains high at 5.35~\textdegree C, indicating large absolute LST errors. This confirms that existing STF methods exhibit substantial degradation on thermal data. DL-based approaches show even larger errors and lower structural consistency, which reflects difficulties in modeling the strong thermal contrasts typical of late-summer conditions. On 30~Sep~2022, EDCSTFN achieves the lowest RMSE of 2.32~\textdegree C and the highest CC, PSNR, and SAM, slightly outperforming ESTARFM in absolute accuracy. ESTARFM and MLFF-GAN follow closely, while STTFN exhibits lower structural and spectral fidelity. Despite EDCSTFN’s better performance, RMSE values above 2~\textdegree C still indicate notable deviations in LST prediction. For 01~Nov~2022, 13~Jun~2023, and 19~Oct~2023, the results clearly indicate that DL-based STF methods outperform the traditional ESTARFM. Across these dates, DL approaches consistently achieve RMSE values below approximately 2~\textdegree C, which is generally considered acceptable for LST estimation. This improvement reflects their superior ability to exploit non-linear spatial and temporal relationships under thermally stable or moderately varying conditions, whereas ESTARFM remains limited by its linear assumptions and shows higher absolute errors. However, this apparent advantage does not necessarily reflect a superior capability to model LST-specific dynamics. Instead, the limited spatio-temporal variability between the target date $t_2$ and the reference pairs ($P_1$ and $P_3$) results in thermally stable conditions, under which all models benefit from reduced temporal complexity. In such cases, DL models are able to effectively capture LST patterns, while their performance gains primarily stem from favorable data conditions. On 28~May~2023, the traditional ESTARFM method outperforms DL-based models. However, the RMSE is above 2~\textdegree C. This suggests that when LST exhibits smoother, quasi-linear temporal evolution, classical STF methods can remain competitive, though they do not provide a substantial accuracy advantage. For 12~Apr~2024 and 19~Sep~2024, all evaluated models fail to accurately reconstruct the spatio-temporal variability of LST. Errors increase substantially, and none of the methods demonstrate consistent spatial or temporal fidelity.

\begin{table}[t]
\caption{Per-date quantitative comparison of STF methods (ESTARFM, STTFN, EDCSTFN, and MLFF-GAN) for LST estimations. Results are reported for multiple target dates using RMSE, SSIM, PSNR, SAM, CC, and ERGAS. For each metric, the best performance per date is highlighted in bold, while arrows indicate whether lower ($\downarrow$) or higher ($\uparrow$) values correspond to better performance.}

\begin{adjustwidth}{-1.5cm}{-1.5cm}
\renewcommand{\arraystretch}{1}
\begin{tabularx}{\fulllength}{l|cccc|cccc}
\toprule
\textbf{Metrics} & \textbf{ESTARFM} & \textbf{STTFN} & \textbf{EDCSTFN} &  \textbf{MLFF-GAN} & \textbf{ESTARFM} & \textbf{STTFN} & \textbf{EDCSTFN} & \textbf{MLFF-GAN}\\

\cmidrule(lr){2-5} \cmidrule(lr){6-9}

 & \multicolumn{4}{c}{\textbf{29 Aug 2022}} & \multicolumn{4}{c}{\textbf{30 Sep 2022}} \\

\midrule

RMSE ($\downarrow$) & \textbf{5.350} & 6.258 & 5.725 & 5.758 & 2.650 & 2.649 & \textbf{2.317} & 2.549 \\

SSIM ($\uparrow$) & \textbf{0.940} & 0.833 & 0.918 & 0.872& \textbf{0.870} & 0.780 & 0.862 & 0.829 \\

PSNR ($\uparrow$) & \textbf{21.410} & 20.048 & 20.821 & 20.772 & 22.900 & 22.894 & \textbf{24.058} & 23.227 \\

SAM ($\downarrow$) & \textbf{8.450} & 9.374 & 8.812 & 9.118 & 6.38 & 7.661 & \textbf{6.336} & 6.798\\

CC ($\uparrow$) &  \textbf{0.640} & 0.537 & 0.600 & 0.537 &  0.650 & 0.536 & \textbf{0.669} & 0.595\\

ERGAS ($\downarrow$)  & \textbf{5.000} & 5.850 & 5.352 & 5.382 & 4.730 & 4.728 & \textbf{4.135} & 4.550\\

\midrule
 & \multicolumn{4}{c}{\textbf{01 Nov 2022}} & \multicolumn{4}{c}{\textbf{28 May 2023}} \\

\midrule
RMSE ($\downarrow$) & 2.720 & 2.137 & 1.579 & \textbf{1.037} & \textbf{2.390} & 3.213 & 2.719 & 3.175 \\

SSIM ($\uparrow$) & \textbf{0.870} & 0.685 &  0.848 & 0.776 & \textbf{0.880} & 0.768 & 0.844 & 0.730 \\

PSNR ($\uparrow$) & 23.190 & 18.980 & 20.449 & \textbf{24.098} & \textbf{23.780} & 21.2 & 22.651 & 21.304 \\

SAM ($\downarrow$) & 6.500 & 4.861 & \textbf{2.937} & 3.666 & \textbf{3.410} & 4.600 & 4.043 & 5.555 \\

CC ($\uparrow$) & \textbf{0.660} &  0.644 & 0.653 & 0.536 & \textbf{0.900} & 0.808 & 0.849 & 0.719 \\  

ERGAS ($\downarrow$)  & 5.65 & 4.445 & 3.283 & \textbf{2.157} & \textbf{2.520} & 3.396 & 2.874 & 3.356 \\

\midrule
 & \multicolumn{4}{c}{\textbf{13 Jun 2023}} & \multicolumn{4}{c}{\textbf{19 Oct 2023}} \\

\midrule
RMSE ($\downarrow$) & 1.950 & 1.993 & \textbf{1.937} & 2.710 & 3.820 & \textbf{1.877} & 2.842 & 3.028\\

SSIM ($\uparrow$) & \textbf{0.900} & 0.817 & 0.892 & 0.843  & 0.820 & 0.845 & \textbf{0.858} & 0.834\\

PSNR ($\uparrow$) & 26.140 & 25.936 & \textbf{26.183} & 23.26& 20.970 & 24.322 & \textbf{24.718} & 20.166\\

SAM ($\downarrow$) & \textbf{3.010} & 3.414 & 3.115 & 3.671 & 8.470 & 5.132 & 5.540 & \textbf{5.131} \\

CC ($\uparrow$) & \textbf{0.910} & 0.867 & 0.890 & 0.846& 0.320 & 0.452 & 0.436 & \textbf{0.467}  \\

ERGAS ($\downarrow$)  &  2.000 & 2.051 & \textbf{1.994} & 2.789& 6.190 & \textbf{3.039} & 4.602 & 4.904 \\

\midrule
 & \multicolumn{4}{c}{\textbf{12 Avr 2024}} & \multicolumn{4}{c}{\textbf{19 Sep 2024}} \\

\midrule
RMSE ($\downarrow$) & 4.480 & 4.365 & 4.186 & \textbf{4.000}& 3.890 & 4.484 & \textbf{3.030} & 3.281 \\

SSIM ($\uparrow$) & 0.800 & 0.789& \textbf{0.827} & 0.806 & 0.800 & 0.759 & \textbf{0.833} & 0.718 \\

PSNR ($\uparrow$) & 20.440 & 20.656 & 21.014 & \textbf{21.414}& \textbf{20.660} & 16.923 & 20.329 & 19.637 \\

SAM ($\downarrow$) & 10.910 & \textbf{8.738} & 9.702 & 9.852& 9.300 & 7.480 & \textbf{7.230} & 7.825 \\

CC ($\uparrow$) & 0.430 & 0.428 & \textbf{0.449} & 0.410& 0.430 & 0.535 & \textbf{0.595} & 0.449 \\

ERGAS ($\downarrow$)  & 6.480 & 6.313 & 6.058 &\textbf{ 5.785}& 5.460 & 6.289 & \textbf{4.249} & 4.601 \\

\bottomrule
\end{tabularx}
\end{adjustwidth}
\label{tab:quantitive}
\end{table}

The average results reported in Table~\ref{tab:quantitive_average} confirm that none of the evaluated STF methods is able to reliably reconstruct LST with high accuracy. Across all methods, RMSE values remain above 3~\textdegree C, indicating substantial absolute temperature errors that are incompatible with many LST-driven applications. Although EDCSTFN achieves the best overall performance among the DL-based approaches, its RMSE of 3.04~\textdegree C remains high. ESTARFM exhibits the weakest performance overall, reflecting the inability of linear, weight-based formulations to capture the non-linear and scale-dependent behavior of LST. Importantly, the relatively small performance gap between traditional and DL-based methods suggests that architectures originally designed for SR fusion do not directly transfer to thermal data.

\begin{table}[h]
\centering
\caption{Average quantitative performance of STF methods (ESTARFM, STTFN, EDCSTFN, and MLFF-GAN) over all test samples. Result are reported using RMSE, SSIM, PSNR, SAM, CC, and ERGAS. For each metric, the best performance per date is highlighted in bold, and $\downarrow$ and $\uparrow$ indicate whether lower or higher values correspond to better performance.}
\renewcommand{\arraystretch}{1}
\begin{tabular}{lcccc}
\toprule
\textbf{Metrics} & \textbf{ESTARFM} & \textbf{STTFN} & \textbf{EDCSTFN} &  \textbf{MLFF-GAN}\\

\cmidrule(lr){2-5}

 & \multicolumn{4}{c}{\textbf{Average}} \\

\midrule

RMSE ($\downarrow$) & 3.406 & 3.372 & \textbf{3.042} & 3.196  \\

SSIM ($\uparrow$) &  0.860 & 0.7845 & \textbf{0.861} & 0.800 \\

PSNR ($\uparrow$) &\textbf{22.436} & 21.371 & 22.279 & 21.736  \\

SAM ($\downarrow$) & 7.054 & 6.408 & \textbf{5.96} & 6.452\\

CC ($\uparrow$) & 0.618 & 0.601 & \textbf{0.643} & 0.576\\

ERGAS ($\downarrow$)  & 4.754 & 4.5139 & \textbf{4.068} & 4.191 \\ 

\bottomrule
\end{tabular}
\label{tab:quantitive_average}
\end{table}

\subsection{Qualitative comparison}
\begin{figure}[H]
\begin{adjustwidth}{-1.5cm}{-1.5cm}
\centering
\includegraphics[width=18cm]{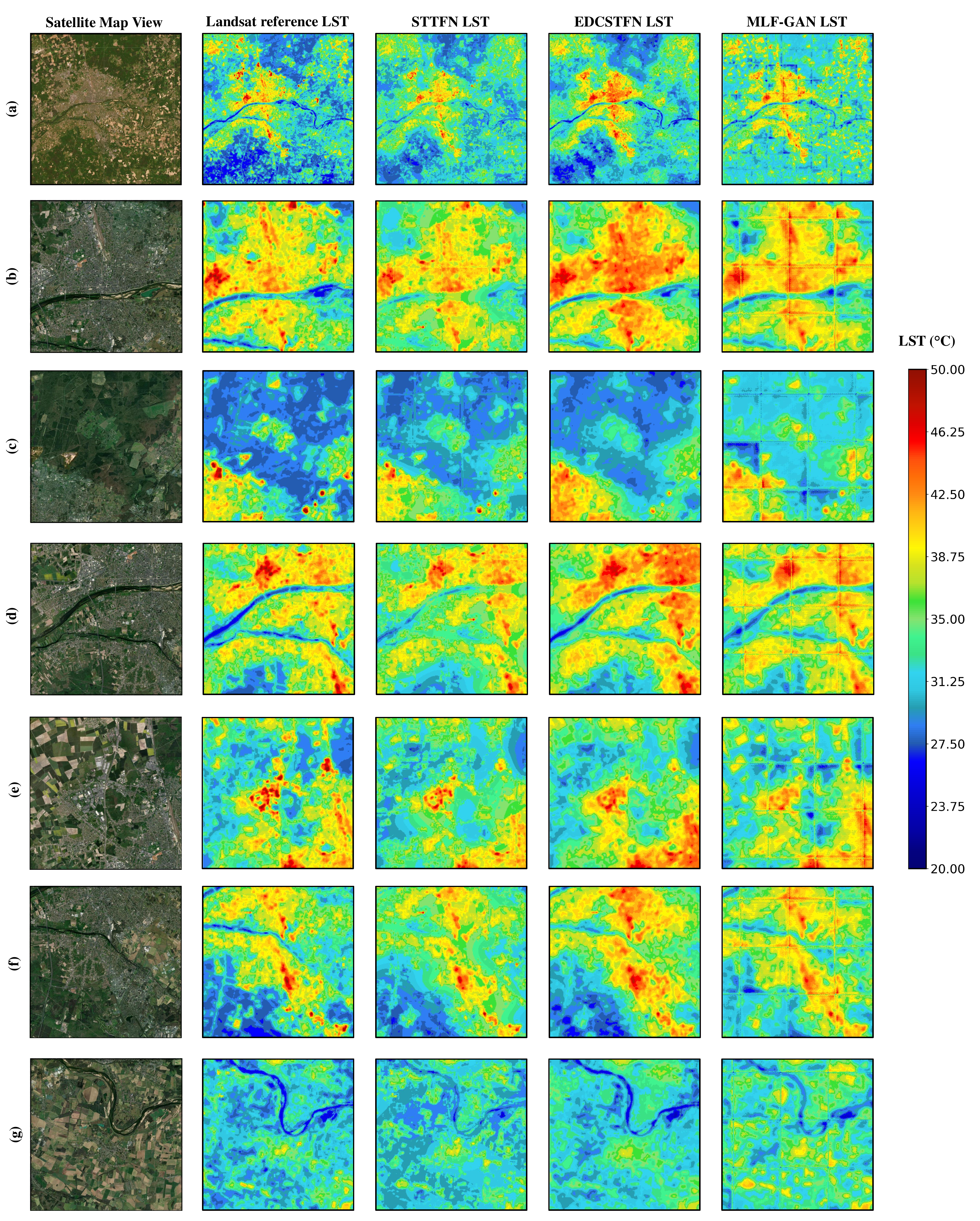} 
\end{adjustwidth}
\caption{Qualitative comparison of DL-based STF methods (STTFN, EDCSTFN, and ESTARFM) over the ROI and seven representative zoomed-in subregions on 13 June 2022: (a) Orléans Métropole, (b) Orléans city center, (c) Orléans forest, (d) a semi-urban corridor along the Loire and Loiret rivers, (e) a major industrial area, (f) a mixed residential and vegetated neighborhood intersected by the Loiret river, and (g) croplands. \label{fig:quality}}
\end{figure}

Figure~\ref{fig:quality} provides a qualitative comparison between the Landsat~8 reference LST and STF-based reconstructions produced by STTFN, EDCSTFN, and MLFF-GAN over the full ROI and seven representative zoomed subregions. Satellite map views are used to associate observed thermal patterns and artifacts with underlying land-cover structures.

In Figure~\ref{fig:quality}(a), corresponding to the full ROI, the Landsat~8 reference LST exhibits a well-defined spatial organization of hot and cold regions driven by land-cover heterogeneity. All STF methods successfully identify large cold areas associated with the Loire river and forested regions. However, substantial discrepancies arise over urban and industrial zones. STTFN systematically underestimates high LST values, leading to muted urban hot spots. In contrast, EDCSTFN overestimates temperatures over built-up areas, producing spatially expanded hot regions that extend beyond actual impervious surfaces. MLFF-GAN yields a more realistic global LST distribution but introduces pronounced block-like artifacts aligned with patch boundaries, which disrupt spatial continuity and degrade physically plausible thermal gradients. 

Figure~\ref{fig:quality}(b) focuses on the Orléans city center, where the reference LST reveals strong thermal heterogeneity linked to dense urban fabric, bridges, and the Loire river, as confirmed by the satellite imagery. EDCSTFN markedly overestimates urban temperatures, generating exaggerated hot zones while partially preserving linear bridge-related structures. STTFN produces overly smoothed temperature fields that suppress fine-scale urban morphology and fail to capture road and bridge-induced thermal contrasts. MLFF-GAN better balances under and overestimation and preserves the primary urban-river contrast. However, severe patch-wise artifacts introduce abrupt and unrealistic transitions between adjacent regions. 

Figure~\ref{fig:quality}(c), corresponding to the Orléans forest, the Landsat~8 reference indicates generally low temperatures with subtle intra-forest variability associated with canopy structure and forest edges. While STTFN and EDCSTFN correctly reproduce the overall cooling effect of dense vegetation, both generate overly homogeneous temperature fields and fail to recover fine-scale thermal variability. Localized warmer areas present in the reference are either smoothed out or merged with surrounding cooler regions. MLFF-GAN introduces artificial spatial patterns unrelated to forest structure, with block-aligned artifacts cutting across natural boundaries and further degrading physical coherence. 

Figure~\ref{fig:quality}(d) depicts the semi-urban corridor along the Loire and Loiret rivers. The reference LST clearly delineates both rivers as cold linear features with sharp banks visible in the satellite imagery. STTFN struggles to preserve narrow water bodies, particularly the Loiret river, resulting in weakened river-induced cooling and blurred boundaries. EDCSTFN more accurately captures river geometry and cooling signals but overestimates adjacent urban temperatures. MLFF-GAN reproduces large-scale thermal patterns but exhibits grid-like artifacts that intersect river courses, disrupting spatial continuity and river morphology. 

Figure~\ref{fig:quality}(e) focuses on a large industrial area characterized by extensive impervious surfaces, dense road networks, and sparse vegetation, as indicated by the satellite map view. The Landsat~8 reference LST exhibits pronounced thermal heterogeneity, with distinct hot spots associated with large industrial buildings and paved surfaces, and cooler patches linked to vegetated buffers. STTFN substantially underestimates LST over this area, resulting in attenuated thermal contrasts that fail to represent the strong heat retention properties of industrial materials. Conversely, EDCSTFN markedly overestimates LST, producing spatially inflated hot regions that extend well beyond the actual industrial footprint visible in the satellite imagery. MLFF-GAN yields more balanced temperature magnitudes. However, its reconstruction is affected by conspicuous block-like artifacts aligned with patch boundaries, which compromise spatial realism. 

Figure~\ref{fig:quality}(f) depicts a heterogeneous residential environment composed of small housing blocks, local road networks, vegetated parcels, and the Loiret river crossing the scene, as confirmed by the satellite map view. The Landsat~8 reference LST reveals fine-scale thermal variability, with warmer residential surfaces interspersed with cooler vegetated areas and a clearly defined cooling corridor associated with the river. STTFN reproduces a relatively realistic overall temperature distribution across residential zones, capturing the general contrast between built-up and vegetated surfaces. However, it fails to preserve the Loiret river’s thermal signature, which becomes largely indistinguishable from adjacent land covers due to excessive spatial smoothing. EDCSTFN substantially overestimates LST across the residential fabric. Although the river-induced cooling effect remains partially visible, it lacks sharp boundaries and physical coherence. MLFF-GAN succeeds in jointly representing both residential thermal patterns and the presence of the Loiret river, but this apparent advantage is undermined by strong grid-like artifacts that introduce artificial spatial structures inconsistent with the underlying urban morphology. 

Figure~\ref{fig:quality}(g) presents an agricultural landscape dominated by croplands with relatively homogeneous land cover, intersected by a curved section of the Loire river, as shown in the satellite map view. The Landsat~8 reference LST is characterized by generally low temperatures and smooth spatial gradients. All STF methods correctly reproduce the overall cooling effect of croplands, indicating that under thermally uniform and slowly varying conditions, the STF task is less challenging. Nevertheless, notable differences persist. STTFN fails to clearly delineate the Loire river, whose cooling signal is substantially weakened and partially blended into surrounding croplands. MLFF-GAN captures the broad thermal patterns and the river-induced cooling effect but introduces artificial block-wise structures that cut across agricultural parcels and natural boundaries.

Overall, the qualitative analysis over the ROI and across urban, industrial, residential, forested, and agricultural subregions reveals systematic limitations shared by DL-based STF methods when applied to LST. Models originally developed for SR-based STF struggle to preserve sharp thermal gradients induced by rivers, bridges, and impervious surfaces, and to accurately reconstruct localized thermal extremes associated with urban and industrial hotspots. STTFN exhibits a pronounced tendency toward spatial oversmoothing, which suppresses fine-scale thermal variability and attenuates narrow cooling features. EDCSTFN consistently overestimates LST over built-up areas, generating spatially inflated hotspots that ignore land-cover boundaries evident in the satellite imagery. MLFF-GAN yields more balanced temperature magnitudes, but introduces strong patch-wise artifacts, leading to artificial spatial discontinuities that undermine physical consistency and thermal realism.

\section{Limitations and future trends}
\label{sec:limitation}
Although substantial advances have been achieved in DL-based STF methods, our experimental analysis demonstrates that directly transferring STF models originally developed for SR to LST remains highly challenging. The intrinsic physical differences between reflectance and thermal signals introduce systematic errors that current DL architectures fail to adequately address. These observations reveal several fundamental limitations of existing STF approaches and motivate the identification of future research directions. This section enumerates the main limitations of current STF-based models when applied to LST data and outlines potential future directions for improvement.

\subsection{Inaccurate LST Estimations}
LST is typically retrieved from satellite observations using physically based algorithms, as described in Section~\ref{sec:lstretrieval}. These retrieval processes introduce uncertainties arising from atmospheric correction, surface emissivity assumptions, sensor noise, and viewing geometry. In the context of STF, multiple satellite-derived LST products are fused. Consequently, inaccuracies present in the input data are directly propagated through the fusion pipeline. When DL-based models are trained on such imperfect inputs, initial LST errors are not corrected but instead may be amplified or accumulated, ultimately degrading the reliability of the fused predictions. This issue complicates the interpretation of STF performance, as observed discrepancies may reflect limitations of the LST retrieval process rather than deficiencies of the STF model itself. For a more realistic and fair evaluation of STF methods, the integration of independent and reliable reference data, such as in situ ground-based temperature measurements, is highly desirable. Validation against such measurements would enable a more accurate assessment of STF model performance while reducing the confounding influence of retrieval-induced LST errors~\cite{krishnan2020intercomparison, shandas2023evaluating}.

\subsection{Cloudy Conditions}
Cloud cover and associated shadows constitute a major challenge for RS applications, as they obstruct the sensor’s ability to acquire clear and temporally consistent observations of the Earth’s surface~\cite{li2019deep}. These effects result in missing or severely corrupted measurements~\cite{mo2021review}. In the context of STF, such gaps are commonly handled through interpolation or gap-filling strategies prior to fusion. However, the accuracy of these approaches remains limited for LST data, especially in regions experiencing persistent cloud cover or rapid land surface dynamics. A more robust strategy consists in explicitly incorporating missing pixels during the training of DL-based STF models. By learning directly from incomplete observations, the network can exploit spatio-temporal dependencies among the available inputs, such as \( X_1(s,t_1) \), \( P_1 \), and \( P_3 \), to infer missing information. This transforms the STF task into a multi-objective learning problem, which aims not only to enhance spatial and temporal resolution but also to reduce or eliminate data gaps in the reconstructed outputs, as formalized in

\noindent Section~\ref{sec:mathematicalformulation}. Recent work, such as MUSTFN~\cite{qin2022mustfn}, demonstrates the potential of this paradigm by producing fused outputs with substantially fewer missing pixels. Nevertheless, the explicit treatment of cloud-induced gaps within STF frameworks remains largely unexplored and constitutes a critical research direction.

\subsection{Poor Generalizability}
When transferring a trained STF model to a new geographical ROI, one of the primary challenges is the domain shift problem. This issue arises because spatio-temporal dynamics, such as seasonal cycles, meteorological conditions, land-cover composition, and urban development, can differ substantially between the training region and the target region. As a result, models trained under specific climatic or environmental conditions may fail to generalize effectively. To mitigate this limitation, fine-tuning becomes essential. Fine-tuning adapts the pretrained model to the target region by updating its parameters using region-specific data, allowing it to capture local temporal dynamics while retaining the transferable representations learned during initial training. Ideally, STF models should be designed to learn robust and generalizable features that remain valid across diverse regions and environmental conditions. For example, studies such as~\cite{huang2024stfdiff} assume that spatial relationships and sensor-related biases are largely invariant across regions, and therefore emphasize adapting temporal representations through fine-tuning.

\subsection{Leveraging Pretrained Models}
The use of pretrained models, although highly effective in many computer vision tasks, remains largely underexplored in STF. To date, \cite{chen2022spatiotemporal} is among the few studies that explicitly investigate the role of pretraining in STF. In broader image fusion and representation learning tasks, large-scale natural image datasets such as ImageNet are commonly used to pretrain deep networks for generic feature extraction. For example, \cite{li2019infrared} and \cite{zhang2021infrared} employed ResNet50 to capture high-frequency features, while \cite{li2018infrared} and \cite{ren2018infrared} adopted VGG19 for deep feature extraction. Similarly, \cite{feng2020fully} utilized DenseNet-201 pretrained on ImageNet to extract hierarchical representations. However, features learned from natural RGB images are not necessarily optimal for STF, particularly for LST, where the underlying physical processes and signal characteristics differ substantially from those of natural imagery. An alternative and potentially more suitable strategy is to pretrain models directly on fusion-oriented datasets, allowing them to learn task-specific representations tailored to multi-source and multi-resolution data. With the rapid growth of fusion datasets in domains such as medical imaging~\cite{li2021medical}, these datasets could serve as a valuable pretraining foundation. Models pretrained in this manner could then be fine-tuned for STF tasks~\cite{azam2022review}.

\subsection{Insufficient Spatial Resolution}
DL-based STF methods for LST are fundamentally constrained by the spatial resolution of the finest available thermal observations, which are typically provided by Landsat at 30~m. While this resolution is adequate for many regional-scale applications, it is insufficient for studies requiring detailed characterization of surface thermal patterns, such as UHI analysis. UHI phenomena are driven by fine-scale urban elements, including roads, buildings, vegetation patches, and urban morphology, that induce strong thermal contrasts at spatial scales smaller than 30~m. As a result, microclimatic variations and localized thermal extremes are often smoothed or entirely missed at this resolution. A promising direction is the integration of higher spatial resolution optical sensors, such as Sentinel-2 (10~m) and PlanetScope (3~m), which provide detailed SR information. Although these platforms lack TIR sensors, their high-resolution optical data can be exploited to guide the spatial disaggregation of coarser thermal observations. Some studies have explored the fusion of MODIS and Sentinel-2~\cite{sanchez2020monitoring} or Landsat~8 and Sentinel-2~\cite{abunnasr2022towards} to estimate LST at 10~m resolution. However, most existing approaches rely on linear assumptions, which limits their ability to capture complex urban thermal processes. Consequently, these methods often struggle to deliver accurate and physically consistent high-resolution LST estimates, particularly in heterogeneous urban environments.

\subsection{Joint spatio-temporal deep learning architectures}
Current STF research largely relies on DL models that prioritize either spatial or temporal modeling, but rarely both in a unified manner. Temporal architectures, such as RNNs and LSTM networks, have demonstrated effectiveness in capturing temporal dynamics in satellite time series. However, they are limited in their ability to explicitly model spatial dependencies and fine-scale spatial heterogeneity. In contrast, spatial architectures, including CNNs, AEs, GANS, and ViT-based models, excel at extracting spatial features and structural patterns, but are not designed to model temporal evolution. Although a small number of studies have explored the integration of CNNs with LSTMs, such as~\cite{meng2022spatio}, fully unified DL-based STF architectures remain largely underexplored in the STF literature. This limitation is particularly critical for LST fusion, where thermal processes are governed by the joint interaction of spatial heterogeneity and temporal dynamics. Developing models that can simultaneously learn spatial structure and temporal evolution explicitly and coherently represents an important direction.

\subsection{Integration of Large Language Models}
Large Language Models (LLMs) have recently demonstrated strong capabilities across a wide range of natural language processing and multimodal reasoning tasks. Recent studies have explored LLMs in diverse domains, including RS data interpretation~\cite{li2024vision}, medical diagnosis~\cite{thirunavukarasu2023large}, and automated code generation~\cite{xu2022systematic}. This rapid progress raises the question of whether LLMs can play a role in STF, not as direct fusion operators, but as complementary components that enrich the fusion process. One potential direction is the extraction of high-level semantic information from satellite imagery and its integration alongside visual features. For example,~\cite{zhao2024image} demonstrated that semantic prompts can be generated from images and subsequently processed by LLMs such as ChatGPT to produce structured textual descriptions. These semantic representations can then be fused with visual features through cross-attention. To date, such LLM-assisted paradigms have not been applied to STF tasks, and no existing methods incorporate textual semantics into the STF of satellite data. Nonetheless, it may offer new opportunities for improving STF, particularly for complex scenes where land-cover context plays an important role. 

\section{Conclusion}
\label{sec:conclusion}
In this work, we have presented a comprehensive survey and experimental analysis of the latest advancements in STF methods for LST estimation, with a particular focus on DL-based approaches. We began by formulating the STF for LST problem mathematically and reviewing the principal DL techniques employed in this domain. Building on this foundation, we proposed a novel taxonomy that categorizes existing methods across key dimensions, including architecture, learning paradigms, training strategies, and the use of pre-trained models. Through extensive experiments on our proposed MODIS-Landsat LST pairs dataset (STF-LST), we systematically assessed the performance and reliability of state-of-the-art STF methods. Our analysis demonstrates that, although DL-based approaches can capture spatio-temporal patterns effectively under conditions of limited variability, directly transferring models designed for SR to LST remains problematic. Key challenges include accurately reconstructing sharp thermal gradients, representing extreme LST values, and handling cloud-induced gaps, all of which limit the practical applicability of current methods. Finally, we identified several promising avenues for future research. By integrating a comprehensive taxonomy with extensive empirical evaluation, this study offers a clear reference for understanding the strengths and limitations of DL-based STF methods, while providing guidance for developing more robust and physically consistent approaches to LST fusion.

\authorcontributions{Conceptualization, S.B., A.H, R.C, and R.N.; methodology, S.B., and A.H.; software, S.B.; validation, S.B., A.H. and R.N.; formal analysis, S.B.; investigation, S.B., A.H, R.C, and R.N.; resources, S.B., A.H, R.C, and R.N.; data curation, S.B., R.N,; writing---original draft preparation, S.B.;
writing---review and editing, S.B., A.H. and R.N;; visualization, S.B.; supervision, A.H, R.C, and R.N.; project administration, S.B., A.H, R.C, and R.N.; funding acquisition, A.H, R.C, and R.N. All authors have read and agreed to the published version of the manuscript.}

\funding{This research was carried out as a part of the CHOISIR project funded by Métropole d'Orléans and Région Centre-Val de Loire.}

\dataavailability{The data and source code supporting the findings of this study are publicly available in the GitHub repository \url{https://github.com/Sofianebouaziz1/STF-LST}. The repository provides a fully reproducible framework for generating the STF-LST dataset, including scripts for downloading, preprocessing, and organising the MODIS and Landsat data pairs using the Google Earth Engine platform. The dataset consists of 51 paired MODIS/Landsat LST images covering the Orléans Métropole area (France) between Apr 2013 and Oct 2024. All satellite data used are publicly accessible from their respective providers. Users must hold a valid Google Earth Engine account to reproduce the dataset.}

\conflictsofinterest{The authors declare no conflicts of interest.} 

\reftitle{References}
\externalbibliography{yes}
\bibliography{references}

\end{document}